\documentclass[a4paper]{article}
\usepackage[english]{babel}
\usepackage[utf8x]{inputenc}
\usepackage[T1]{fontenc}
\usepackage{times}
\usepackage{pgf,tikz,pgfplots}
\usepackage{mathrsfs}
\usepackage{tabularx}
\usepackage[round]{natbib}
\usepackage{amsmath}
\usepackage{algorithm}
\usepackage{algpseudocode}
\usepackage{amsfonts} 
\usepackage{soul}
\usepackage{bm}
\usepackage{amssymb}
\usepackage{multirow}
\usepackage{float}
\usepackage{pdflscape}
\usepackage{afterpage}
\usepackage[colorinlistoftodos]{todonotes}
\usepackage{subcaption}
\usepackage{caption}
\usepackage{booktabs,makecell}
\usepackage{graphicx}
\usepackage{csvsimple}
\usepackage{xcolor}
\usepackage{capt-of}
\usepackage[colorlinks=true, allcolors=blue]{hyperref}
\usepackage{fixmath}
\usepackage{forest}
\pgfplotsset{compat=1.14}
\usepackage{array}
\newcolumntype{?}{!{\vrule width 1pt}}
\usepackage{arydshln}

\usetikzlibrary{arrows.meta,
                positioning, calc, 
                shapes,chains, arrows, trees,patterns}

\newcommand\blfootnote[1]{%
  \begingroup
  \renewcommand\thefootnote{}\footnote{#1}%
  \addtocounter{footnote}{-1}%
  \endgroup
}

\definecolor{uququq}{rgb}{0.25098039215686274,0.25098039215686274,0.25098039215686274}
\tikzset{
    quote/.style={{|[width=2mm]}-{|[width=2mm]}}
}

\newtheorem{theorem}{Claim}
\newtheorem{corollary}{Assumption}

\tikzset{%
  every neuron/.style={
    circle,
    draw,
    minimum size=1cm
  },
  neuron missing/.style={
    draw=none, 
    scale=4,
    text height=0.333cm,
    execute at begin node=\color{black}$\vdots$
  },
}

\newif\ifshowcomments

\newcommand{\ef}[1]{\ifshowcomments{\color{black}#1}\fi}
\newcommand{\greta}[1]{\ifshowcomments{\color{black}#1}\fi}
\newcommand{\gl}[1]{\ifshowcomments{\color{black}#1}\fi}

\newcommand{\efsec}[1]{\ifshowcomments{\color{black}#1}\fi}

\date{}
\title{Periodic Freight Demand \greta{Estimation} for Large-scale Tactical Planning}
\author{Greta Laage\textsuperscript{a,b,}\footnote{Corresponding author.}, Emma Frejinger\textsuperscript{c,b}, Gilles Savard\textsuperscript{a,d}}
\begin{document}

\showcommentstrue

\definecolor{sexdts}{rgb}{0.1803921568627451,0.49019607843137253,0.19607843137254902}
\definecolor{dbwrru}{rgb}{0.8588235294117647,0.3803921568627451,0.0784313725490196}
\definecolor{rvwvcq}{rgb}{0.08235294117647059,0.396078431372549,0.7529411764705882}
\maketitle

\blfootnote{E-mail addresses: greta.laage@polymtl.ca (G. Laage)} %emma.frejinger@umontreal.ca (E. Frejinger), %gilles.savard@polymtl.ca (G. Savard)}

\noindent
\textsuperscript{a} Department of Mathematics and Industrial Engineering, \'Ecole Polytechnique de Montr\'eal%, 2500 Chemin de Polytechnique, Montr\'eal QC H3T 1J4 Canada
\\
\textsuperscript{b} CIRRELT%, 2920 Ch de la Tour, Montr\'eal QC H3T 1J4, Canada
\\
\textsuperscript{c} Department of Computer Science and Operations Research, Universit\'e de Montr\'eal%, 3150 Rue Jean-Brillant, Montr\'eal QC H3T 1N8, Canada
\\
\textsuperscript{d} IVADO%, 6666 Rue Saint-Urbain 4e étage, suite 480, Montr\'eal QC H2S 3H1, Canada

%Declaration of Interest: None.

%\tableofcontents

\begin{abstract}

\ef{Freight carriers rely on tactical planning to design their service network to satisfy demand in a cost-effective way. For computational tractability, deterministic and cyclic Service Network Design (SND) formulations are used to solve large-scale problems.} A central input is the \emph{periodic demand}, that is, the demand expected to repeat in every period in the planning horizon. \ef{In practice, demand is predicted by a time series forecasting model and the periodic demand is the average of those forecasts. This is, however, only one of many possible mappings. The problem consisting in selecting this mapping has hitherto been overlooked in the literature. We propose to use the structure of the downstream decision-making problem to select a good mapping. For this purpose, we introduce a multilevel mathematical programming formulation that explicitly links the time series forecasts to the SND problem of interest. The solution is a periodic demand estimate that minimizes costs over the tactical planning horizon.}
We report results in an extensive empirical study of a large-scale application from the Canadian National Railway Company. 
\ef{They} clearly show the importance of the periodic demand estimation problem. Indeed, the planning costs exhibit an important variation over different periodic demand estimates and using an estimate different from the mean forecast can lead to substantial cost reductions. Moreover, the costs associated with the \gl{periodic} demand estimates based on forecasts were comparable to, or even better than those obtained using the mean of \emph{actual} demand.

\end{abstract} 

\paragraph{Keywords}  Freight transportation, tactical planning, large-scale, periodic demand, forecasting  demand. 

% Nouvelle version.
\section{Introduction}
Freight transportation is essential to society and its economic development. In order to satisfy demand in a cost-effective way, freight carriers are faced with a multitude of planning problems. In this context, Service Network Design (SND) is an important class of problems. Consider, for example, the Multicommodity Capacitated Fixed-charge Network Design (MCND) problem \citep{magnanti1984network}. The objective is to design a capacitated network -- a tactical plan -- \greta{allowing} to transport demand for a set of commodities between different origin-destination pairs at a minimum cost. The latter is given by the sum of fixed and variable costs. The tactical plan is defined over a given period (e.g., a week) and is repeated over a planning horizon (e.g., a few months). Given this cyclic nature of the tactical plan, it relies on an accurate representation of \textit{periodic demand}\efsec{, i.e., demand that repeats in each period}.

In any realistic setting, demand for commodities is subject to uncertainty. This has naturally led to stochastic SND formulations \citep[e.g.,][]{crainic2020}. As even deterministic SND problems are NP-hard, stochastic formulations are limited to fairly small size problems and cannot yet be used in most real large-scale applications. Hence, a wealth of applications \ef{in practice} relies on deterministic formulations and point estimates of periodic demand. \greta{ We focus on the problem of defining such periodic demand for large-scale deterministic SND formulations. It has been overlooked in the literature since most of the work on SND focus on the optimization problem only (demand is assumed to be fixed and known in numerical studies). In practice, demand is typically predicted by a time series forecasting model and the periodic demand is the average of those forecasts.} 

The key motivation behind our work resides in the mapping from time series forecasts to periodic demand. Indeed, the average is merely one, out of many possible mappings. If the SND problem is sensitive to extreme demand values, selecting another mapping than the average, such as the third quartile, \efsec{could} lead to cost reductions.
Despite its \ef{potential} importance, there is no study in the literature focusing on the periodic demand estimation problem linking time series forecasts to the tactical planning problem of interest. Our work addresses this gap, and we use a MCND formulation for illustration purposes.

The impact of demand forecast errors on revenue has been studied in the context of airline revenue management. Through simulation analysis in a simplified setting, \cite{weatherford_belobaba2002} show that reducing demand forecast errors by 25\% increase revenue by a minimum of 1-2\%. 
\cite{fiig2019} confirm those findings in a more complex airline revenue management setting and show that reduced forecast errors lead to increased revenue. As opposed to passenger transportation, freight carriers typically have flexibility regarding the routing of demand as long as it respects certain constraints, such as delivery time. Moreover, the freight demand origin-destination matrices are often unbalanced, meaning that there can be excess supply in certain directions. It implies that the cost associated with demand forecast errors can vary over commodities. This further motivates the importance of linking the periodic demand estimation and the corresponding SND problem.

Our proposed methodology \efsec{to address the periodic demand estimation problem} proceeds in two steps: First we forecast demand for a given set of commodities for \textit{each period} of the planning horizon. This corresponds to a multivariate multistep time series forecasting problem. \efsec{In the second step, we solve a mathematical program whose solution is the periodic demand that minimizes fixed and variable costs over the planning horizon. It is based on mappings} from the time series forecasts to periodic demand. \ef{D}ifferent mappings \efsec{correspond} to different periodic demand estimates. We propose to use the structure of the downstream SND problem when selecting the mapping. \efsec{The proposed} mathematical program \efsec{hence} explicitly links the mappings to the \efsec{SND formulation (in our case MCND)}. %It selects the periodic demand that minimizes the fixed and variable costs over the planning horizon.

\subsection{\greta{Review of Work Related to Freight Demand Forecasting}}

The multivariate multistep time series forecasting problem in the first step of our meth\-odology is particularly challenging for \efsec{three main} reasons. First, \ef{forecasting models rely on} historical data that result \ef{from} operational decisions. \ef{Hence, the demand data are} constrained by available capacity. \ef{This means that demand recorded in the data may not correspond to actual demand. In the literature this is referred to as truncated or censored data depending on how the data are constrained by the supply.} Second, there are a large number of commodities and \greta{a} relatively long forecasting horizon which can lead to spatiotemporal correlations. Third, the demand varies over time and long-term dependencies and seasonality can be specific to each commodity. This highlights the potential need for modeling both commodity specific behavior and correlations between commodities. 
\ef{There is an extensive body of literature on censored / truncated data and time series forecasting. In this work we do not propose methodological contributions to this literature. Rather, we use existing models to analyze the performance of our periodic demand estimation methodology. For this purpose\gl{,} we provide next a brief and high-level view of the literature where we first  discuss censored and truncated data, followed by time series forecasting.}

\ef{Truncated or censored data arise in a broad variety of applications in different disciplines (e.g., engineering, sociology and economics). In our case, truncated data refer to demand records that are missing, for example, due to a transport service being unavailable. On the contrary, censored data refer to partially observed demand, for example, when the capacity of a given service is reached. Ignoring related issues may result in positively or negatively biased forecasts depending on the type of censoring or truncation \citep[see, e.g.,][for an in-depth discussion of these issues and an overview of methods with application in revenue management]{Queenan2007,vulcano2012estimating,azadeh2014taxonomy,Weatherford2016}.}

\ef{The literature identifies essentially two broad ways to deal with censored or truncated data \citep{Queenan2007}. First, adapt the data capture process. This can consist in capturing additional data (e.g., lost sales) or to aggregate data records such that they are no longer censored or truncated. In this paper we adopt an aggregation approach and we provide a motivation for this choice in Section~\ref{sec:meth_forecasting}. The second and more sophisticated approach consists in defining a statistical method whose aim is to recover the latent -- unobserved -- distribution generating the observations.  Many of the existing methods are rooted in statistical survival analysis \citep[e.g.,][]{KleiMoes05}. Unfortunately, standard methods are based on independence assumptions that typically do not hold in a transportation setting due to its intrinsic challenges \citep{FieldsEtAl21Trunc}. Namely, (i) censoring/truncation is common (ii) demand for different commodities impact each other as they share the same capacity (iii) alternative transport services (substitutes) are common which result in so-called \emph{recapture} (demand for second-best options) and \emph{spill} (demand is lost to competition). Because of these challenges, specialized methods are required. \cite{FieldsEtAl21Trunc} propose such a method and they also provide an excellent literature review with a focus on transport applications.}

We now turn our attention to the extensive body of literature in statistics and machine learning \efsec{dealing with} time series forecasting.
\greta{Statistical methods have focused on parametric models informed by domain expertise, such as autoregressive processes, moving averages \citep{box2015time}, and exponential smoothing \citep{gardner1985exponential}. Methods have been developed for univariate seasonal time series, such as seasonal ARIMA \citep{box2015time} and Holt Winter's method \citep{holt2004forecasting, winters1960forecasting}. Vector autoregressive (VAR) models extend the autoregressive (AR) models to multivariate time series prediction problems \citep{lutkepohl2013introduction}. They capture interrelationships among multiple variables, and Vector Error-Correction (VEC) models are preferably used when the time series are cointegrated. However, the number of parameters in VAR and VEC models is large for problems with many variables. This can lead to uncertainty in the estimation of the parameters which, in turn, may lead to inaccurate predictions \citep{litterman1986forecasting}. 
More recently, the machine learning \citep{ahmed2010empirical} and deep learning \citep{Goodfellow2016} literature have provided non-parametric models that learn temporal dynamics in a data-driven manner.} The capacity of neural networks to model complex data to forecast, for instance, traffic flows is increasingly exploited \citep{nguyen2018deep}. The Long Short-Term Memory (LSTM) recurrent neural network \citep{hochreiter1997long, sutskever2014} is a successful architecture to model both short-term and long-term dependencies \citep{Langkvist2014}. Nevertheless, empirical evidence \efsec{suggests} that it is still hard to achieve a level of accuracy comparable to that of classic time series models \citep[e.g.,][]{makridakis2018statistical} \efsec{often due to limited data quantity}.

\greta{Both models from machine learning and statistics have been used and compared for a variety of applications, including transportation \citep{Karlaftis2011}. In this context we distinguish studies focusing on one particular carrier (our setting) from those studying freight flow from a transport system perspective. In the latter case, they often consider a strategic setting requiring long-term demand forecasts \citep[e.g.,][]{Jong2004,Cambridge2008,Chow2010}.
The literature focusing on a specific freight carrier and, in particular, on intermodal transportation is scarce. A few works have, however, focused on forecasting intermodal demand over small networks of port terminals \citep{milenkovic2017container} with either statistical models \citep{schulze2009forecasting} or neural networks \citep{tsai2017using}.}

\subsection{Contributions}

\ef{This} paper offers both methodological and empirical contributions. First, we formally introduce the periodic demand estimation problem and propose a two-step methodology. \ef{The core methodological contribution resides in the problem definition and the formulation used in the second step. More precisely, b}ased on time series forecasts obtained in the first step, we propose a multilevel mathematical programming formulation whose solution is a periodic demand estimate that minimizes fixed and variable costs. The formulation hence explicitly links the periodic demand estimates to the tactical planning problem of interest. \efsec{Under weak assumptions, i}t is computationally tractable as it can be solved sequentially to optimality. 

Second, we describe a real large-scale application \greta{for intermodal (container) transportation at the} Canadian National Railway Company (CN) -- one of the largest railroads in North America. \efsec{We use the MCND formulation proposed by \cite{Morganti2019} for a block planning problem. They focused on the SND problem -- supply optimization -- and treated demand as fixed and known using historical values. Unlike them, we focus on periodic demand estimates obtained from forecasts, i.e., the input to the MCND formulation. The methodological and empirical contributions are therefore distinct from those of \cite{Morganti2019}.}
We present an extensive empirical study clearly showing the importance of the periodic demand estimation problem. \ef{To assess the sensitivity to different forecasting models and related error distributions}, we compare forecasting models from the statistics and deep learning literature. In turn\ef{,} we analyze the impact of the definition of periodic demand distinguishing between time series forecast errors and the errors introduced by different periodic demand estimates. Moreover, we benchmark against \ef{the} approach used in practice that consists in averaging time series forecasts.  

\subsection{Paper Organization}

The remainder paper is structured as follows. Next\gl{,} we formally introduce the periodic demand estimation problem. In Section~\ref{section:formulation_P}\efsec{,} we describe the proposed two-step methodology. We then focus on empirical results, first introducing our application in Section~\ref{section:block_plan}, followed by the results in Section~\ref{section:results}. Finally, Section~\ref{section:ccl} concludes and outlines some directions for future research.

\section{Problem Description}
\label{section:pb_dest}

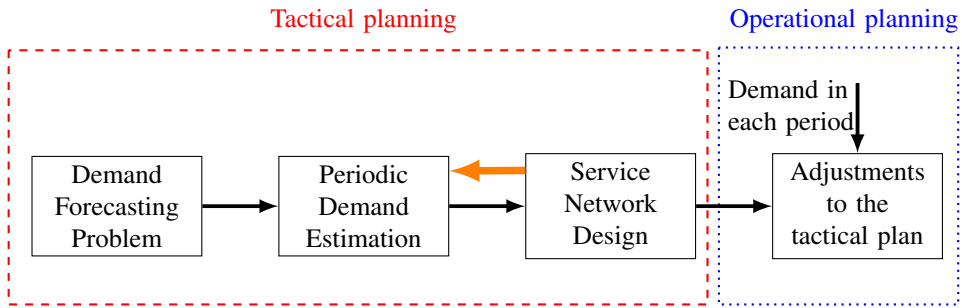
\begin{figure}[htbp]
\centering
\begin{tikzpicture}[box/.style={rectangle,draw}]
% Place nodes
\node (n0) [box, text width=2cm,align=center]  {Demand \\ Forecasting \\ Problem};
\node (n1) [box, text width=2cm,align=center,right=of n0,node distance = 12em]  {Periodic \\Demand \\Estimation};
\node (n2) [box, text width=2cm,right=of n1, node distance = 12em,align=center] {Service \\Network \\Design};  
\node (n3) [box, text width=2cm,right=of n2, node distance = 12em,align=center] {Adjustments \\to the \\tactical plan};
\node [above of = n3, node distance = 5em] (nothing) {};

\node (nothing2) [left of = n3, node distance = 2em] {};
\node (n4) [text width=2cm, above of = nothing2, node distance = 3.7em] {Demand in \\ each period};

\node (n5) [text width=5cm, above left=4.2em and -13em of n3] {\textcolor{blue}{Operational planning}};

\draw[red,thick,dashed] ($(n0.north west)+(-0.3,1.4)$)  rectangle ($(n2.south east)+(0.15,-0.6)$) {};
\node (nothing3) [text width = 5cm, align=center, above of = n1, node distance = 7em, text = red] {Tactical planning};

\draw[blue,thick,dotted] ($(n3.north west)+(-0.7,1.4)$)  rectangle ($(n3.south east)+(0.2,-0.6)$) {};

\draw[>=latex,->, line width=0.5mm] (n0) -- (n1);
\draw[>=latex,->, line width=0.5mm] (n1) -- (n2);
\draw[>=latex,->, line width=0.5mm] (n2) -- (n3);
\draw[>=latex,->, line width=0.5mm] (nothing) -- (n3) ;
\draw[>=latex,<-, line width = 1mm,color=orange] ($(n1.north east)+(0,-0.2)$) -- ($(n2.north west)+(0,-0.23)$);

\end{tikzpicture}
\caption{\greta{Tactical and operational planning processes of a freight carrier}}
\label{fig:planning_processes}
\end{figure}

We start by briefly summarizing the planning process we consider \greta{(see illustration in Figure~\ref{fig:planning_processes}), taking} the point of view of a freight carrier \ef{wishing} to define a tactical plan. First, the carrier estimates the periodic demand for each commodity over the tactical planning horizon. Second, the periodic demand estimates are used as an input to solve the tactical planning problem of interest. The latter involves design decisions that are fixed over the tactical planning horizon, and flow decisions \ef{in each period}. \greta{The tactical level is framed with a red, dashed, line in Figure~\ref{fig:planning_processes}}. %that can be adjusted in each period.
At the operational level \greta{(framed with a blue, dotted, line in the figure)} the tactical plan is adjusted -- i.e., the flow decisions -- according to actual demand realizations. In addition to these adjustments, other decisions could be taken to cope with demand fluctuations, such as outsourcing. There are hence two sources of costs to consider in the tactical planning process: the fixed cost of the tactical plan (design decisions) and the variable cost (flow and outsourcing decisions) resulting from the adjustments in each period. Our objective, illustrated \efsec{in the figure} with a thick orange left arrow, is to use the information from the planning problem to estimate the periodic demand that minimizes the tactical costs.

We attend to large-scale problems that require a deterministic formulation \greta{of the SND problem} to be \efsec{computationally} tractable. Therefore, at the tactical planning level, demand is treated as fixed and known while, in reality, it varies in each period. In this section we describe in detail the problem of estimating the periodic demand so as to minimize fixed and variable costs. We first introduce notation related to \greta{the} tactical and operational planning time horizons. We then define the various concepts of demand we encounter, followed by a description of an MCND formulation. 
Finally, we describe links between observed demand, periodic demand and the MCND formulation, which formally introduces our problem.

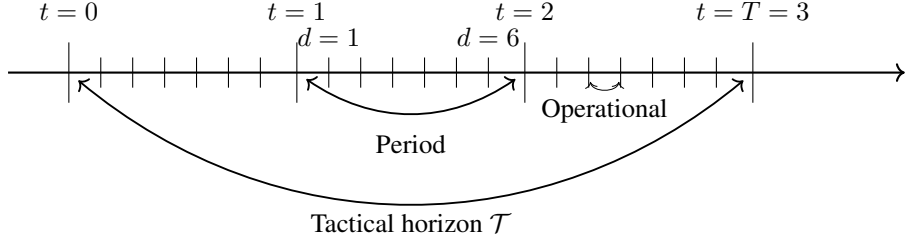
\begin{figure}[!htbp]
	\centering
	\begin{tikzpicture}
	\draw[thick, ->] (0.2,0) -- (12,0) node [below] {};
	
	\foreach \x in {1,...,4}
	\draw (3*\x-2, 0.4) -- node[pos=0.5] (point\x) {} (3*\x-2, -0.4);
	
	\foreach \x in {1,...,6}
	\draw (3+3*\x*0.14-2, 0.2) -- node[pos=0.5] (pointD\x) {} (3+3*\x*0.14-2, -0.2);
	
	\foreach \x in {1,...,6}
	\draw (6+3*\x*0.14-2, 0.2) -- node[pos=0.5] (pointE\x) {} (6+3*\x*0.14-2, -0.2);
	
	\foreach \x in {1,...,6}
	\draw (9+3*\x*0.14-2, 0.2) -- node[pos=0.5] (pointF\x) {} (9+3*\x*0.14-2, -0.2);
	
	\path (point1) node [above = 0.45cm of point1] {$t=0$};
	\path (point2) node [above = 0.45cm of point2] {$t=1$};
	\path (point3) node [above = 0.45cm of point3] {$t=2$};
	\path  (point4) node [above = 0.45cm of point4] {$t=T=3$};
	
	\path (pointE1) node [ above = 0.10cm of pointE1] {$d=1$};
	\path (pointE6) node [ above = 0.10cm of pointE6] {$d=6$};

	\path (pointF2) node [black, above = 0.10cm of pointF2] {};
	\path (pointF3) node [black, above = 0.10cm of pointF3] {};

	\path  (point2) edge[bend right = 35, <->, line width=0.25mm] node [below = 0.15cm of point1] {Period} (point3) ;
	
	\path  (point1) edge[bend right = 40, <->, line width=0.25mm] node [anchor=north] {Tactical horizon $\mathcal{T}$} (point4) ;
	
	 \path  (pointF2) edge[bend right = 80, <->, line width=0.10mm] node [anchor=north] {Operational} (pointF3) ;
	
	\end{tikzpicture}
	\caption{Time scales for planning of a freight carrier}
	\label{fig:time_rep}
\end{figure}

For the demand forecasting problem, it is important to distinguish the tactical and operational planning horizons. We therefore introduce two different notations related to time. First, the tactical planning horizon $\mathcal{T}$ can be divided into periods of equal length $t = 1, \ldots, T$. Second, each period $t$ can be further divided into $D$ time periods, $d = 1, \ldots,D$, that we here refer to as the operational horizon. Figure~\ref{fig:time_rep} provides an illustration where the tactical horizon is composed of \efsec{$T = 3$} weeks, and a week $t$ is composed of $D = 7$ days. 

Let $\mathbf{y}_t$ be the demand vector of period $t$, $\mathbf{y}_t = ( y_{t1}, \dots, y_{tK})^{\top}$ where $y_{tk}$ is the quantity of commodity $k$ to be transported during period $t$. In this context, a commodity $k$ is characterized by its origin $o_k$, destination $d_k$ and type $\gamma_k$. We denote the set of commodities $\mathcal{K}$ and its cardinality $K$.
Let $\mathbf{y}_d^t$ be the demand vector for each operational time $d=1,\ldots,D$, within period $t$, $\mathbf{y}_d^t = ( y_{d1}^t, \dots, y^t_{dK})^{\top}$ where $y_{dk}^t$ is the demand for commodity $k$ to be carried at time $d$ in period $t$. The demand for a period $t$ is hence 
\begin{equation}
    y_{tk} = \sum_{d =1}^D y_{dk}^t, \quad k \in \mathcal{K}.
    \label{eq:demand_relation}
\end{equation}
\efsec{Furthermore, w}e introduce the demand matrix for horizon $\mathcal{T}$, $\mathbf{Y}^{\mathcal{T}} \in \mathbb{R}_+^{T \times K}$, with $[\mathbf{Y}^{\mathcal{T}}]_{tk} = y_{tk}$.
For a given tactical planning horizon $\mathcal{T}$, the plan is repeated at each $t = 1, \ldots, T$. 
Let $\mathbf{y}^{\text{p}\mathcal{T}}$ be the periodic demand vector for tactical horizon $\mathcal{T}$,  $\mathbf{y}^{\text{p}\mathcal{T}} = ( y_1^{\text{p}\mathcal{T}}, \dots, y_K^{\text{p}\mathcal{T}})^{\top}$, where $y_k^{\text{p} \mathcal{T}}$ is the periodic demand for commodity $k$. To simplify the notation, we henceforth remove the superscript $\mathcal{T}$ but recall that the periodic demand and the demand matrix always refer to a given horizon. 

We focus on estimating the periodic demand $\mathbf{y}^{\text{p}}$.
In this context it is important to note that time series forecasting models produce demand forecasts for each commodity in \emph{each period} $t$. That is, at period $t_0$, the forecasting models output an estimate of $\mathbf{Y}$ denoted $\hat{\mathbf{Y}}$ which consists in $T$ point estimates $\hat{\mathbf{y}}_{t_0+1}, \dots, \hat{\mathbf{y}}_{t_0+T}$. 
The periodic demands $y_k^{\text{p}}$ are then estimated from these forecasts $\hat{\mathbf{Y}}$, or, for validation or analysis, from $\mathbf{Y}$. Let $h$ denote the mapping of $\mathbf{Y}$ to a periodic demand vector $\mathbf{y}^{\text{p}}$:
\begin{equation}
    \begin{split}
    h \colon \quad \mathbb{R}_+^{T \times K} & \to \mathbb{R}_+^{K}  \\
    \mathbf{Y} &\mapsto \mathbf{y}^{\text{p}} = h(\mathbf{Y}).
    \end{split}
    \label{eq:h_mapping}
\end{equation}
When the periodic demand is a mapping of the forecasts, we use the notation $\hat{\mathbf{y}}^{\text{p}} = h(\hat{\mathbf{Y}})$. Our objective is to define the mapping $h$ minimizing fixed and variable planning costs.

\begin{figure}[htbp]
\centering
  \includegraphics[width=\linewidth]{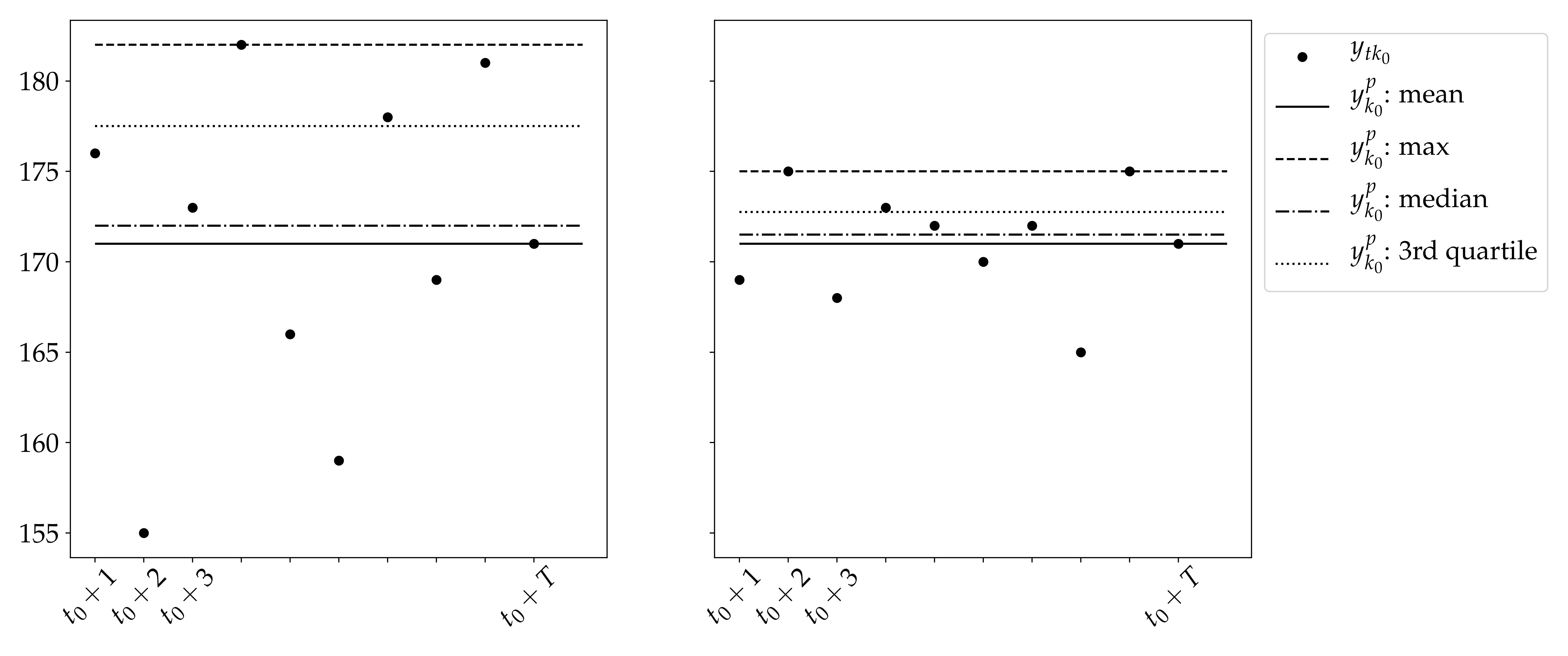}
  \caption{Illustration of a periodic demand from point estimates for $T$ periods}
  \label{fig:explain_periodic}
\end{figure}

\ef{In the following we provide a more detailed motivation of our work by means of a small example.}
Figure~\ref{fig:explain_periodic} \ef{has} two graphs showing two demand distributions (one dot per period) with identical mean (depicted with a solid line). The mean represents one particular mapping $\mathbf{y}^{\text{p}} = h(\mathbf{Y})$. \ef{However, there are many other possible mappings.} In the example, we illustrate three \ef{of those}: the maximum, median and third quartile. As opposed to the mean, these other mappings do not result in the same periodic demand estimates in the right\greta{-} and left-hand graphs. If the SND problem is sensitive to extreme values, then one can clearly see that the third quartile or the maximum \efsec{could} be more adequate mappings than the mean.
The importance of finding an appropriate mapping is the key motivation behind our work and we use the structure of the decision-making problem of interest (SND) to guide the choice of mapping.

We now introduce a path-based MCND formulation \citep{crainic2000service} that we use for illustrating our methodology. An arc-based formulation can be found, e.g., in \cite{chouman2017commodity}. Let $\mathcal{G} = (\mathcal{N}, \mathcal{A})$ denote a space-time graph where $\mathcal{N}$ is the set of nodes and $\mathcal{A}$ is the set of arcs. 
Commodity $k$ uses a path $p$, i.e., a sequence of arcs in $\mathcal{G}$. The source node of the first arc is $o_k$, and the sink node of the last arc is $d_k$. Let $\mathcal{P}$ denote the set of paths. In the case of insufficient capacity, demand is outsourced and we denote $\mathcal{P}^{\text{out}}\subset \mathcal{P}$ the paths corresponding to outsourcing options. % such that $\mathcal{P}^{\text{out}} \subset \mathcal{P}$. 
Furthermore, let $\mathcal{P}_k$ denote the set of paths for commodity $k$, $\mathcal{P}_k^{\text{out}} \subset \mathcal{P}_k$ the outsourcing paths for commodity $k$ 
%such that $\mathcal{P}_k^{\text{out}} \subset \mathcal{P}_k$ 
and $\mathcal{K}_p$ the set of commodities that can use $p$. Note that here we refer to outsourcing in a broad sense. It could mean outsourcing to a third party, or making use of additional capacity from the same carrier that was not originally part of the plan. For example, in our intermodal rail transportation application (Section~\ref{section:block_plan}), outsourcing means using capacity from non-intermodal trains.

The MCND problem consists in satisfying demand at minimum cost. It has two categories of decision variables: Binary design variables $z_p, ~\forall p\in \mathcal{P},$ equal to one if path $p$ is used and zero otherwise, and flow variables $x_{pk} \geq 0, ~\forall k\in \mathcal{K}, p\in \mathcal{P}_k$. Depending on the type of freight (bulk versus containers, for instance), $x_{pk}$ \efsec{could be} integer. The path-based mixed integer linear programming formulation is:  
\begin{flalign}
    \textbf{MCND} & \hspace{0.5cm} \min_{z, x}  \sum_{p \in \mathcal{P}} C_p^{\text{design}} z_p   +  \sum_{k \in \mathcal{K}} \sum_{p \in \mathcal{P}_k \setminus \mathcal{P}_k^{\text{out}}} C_p^{\text{flow}} x_{pk} +  \sum_{k \in \mathcal{K}} \sum_{p \in \mathcal{P}_k^{\text{out}}}  C_p^{\text{out}} x_{pk} \label{eq:mcnd_objective} \\
    & \hspace{0.6cm} \text{s.t.} \hspace{0.45cm} \sum_{p \in \mathcal{P}_k} x_{pk} = y_k^{\text{p}}, \hspace{3.9cm} k \in \mathcal{K}, \label{eq:mcnd_demand}\\ 
    & \hspace{1.4cm} \sum_{k \in \mathcal{K}_p} x_{pk} \leq u_p z_p, \hspace{3.6cm} p \in \mathcal{P}, \label{eq:mcnd_cap}\\ 
    & \hspace{1.4cm} x_{pk} \geq 0 , \hspace{4.9cm} k \in \mathcal{K}, p \in \mathcal{P}_k, \label{eq:mcnd_flowPos}\\ 
    & \hspace{1.4cm} z_p \in \{0,1\}, \hspace{4.45cm}  p \in \mathcal{P}.  \label{eq:mcnd_design}
\end{flalign}
The objective function \eqref{eq:mcnd_objective} includes a fixed design cost 
$C_p^{\text{design}}\geq0$ for the paths built to transport demand. The second cost is the variable flow cost $C_p^{\text{flow}}\geq0$ which accounts for satisfied demand and the third cost \efsec{$C_p^{\text{out}}\geq C_p^{\text{flow}}$} is the flow cost of outsourced demand. \efsec{Although not strictly necessary, we separate the latter two terms for the analysis of the results in Section~\ref{section:results}.} Constraints \eqref{eq:mcnd_demand} ensure that the periodic demand is satisfied for each commodity. Constraints \eqref{eq:mcnd_cap} enforce flows on selected paths only, and that the flow does not exceed the path capacity, $u_p$.

\textbf{MCND} is solved to obtain a tactical plan based on a given periodic demand. However, in practice, demand varies from one period to another. The tactical plan is therefore adjusted at the operational planning level. That is, in each period the commodity flows can be adjusted to satisfy the actual demand value of this period also taking into account other uncertain aspects, such as schedule delays. The observed data $\mathbf{y}_d^t$, typically used for training forecasting models, result from this operational planning process. Consequently, data at this level of detail can be constrained by the available services and the observed demand may therefore \efsec{be censored or truncated.}
%not correspond to the true demand. This is known as censored data in the literature \citep[e.g.,][]{ParkEtAl07}.

In summary, we focus on estimating periodic demand $\mathbf{y}^{\text{p}}$ at a given time $t_0$ for a tactical planning horizon $\mathcal{T}$. The demand forecasts $\hat{\mathbf{y}}_{t_0+1},\ldots,\hat{\mathbf{y}}_{t_0+T}$ are obtained using historical data of demand $\{\mathbf{y}_d^s, s=t_0-1,t_0-2,\ldots,t_0-H, d=1,\ldots,D_s\}$ where $H$ is the number of periods in the historical data and $D_s$ is the number of operational time intervals in each period $s$. 
The periodic demand should be defined such that it minimizes fixed costs, as well as variable costs associated with adapting the plan over the tactical planning horizon. It is hence necessary to link the mapping $h$ to the tactical planning problem of interest. In the following section, we propose a formulation for this purpose using \textbf{MCND} as an example of tactical planning problem formulation.

\section{Periodic Demand Estimation}
\label{section:formulation_P}

Each time a tactical plan is to be defined, our approach proceeds in two steps. First, we use a time series forecasting model to predict demand for each period 
in the planning horizon. Second, we solve a multilevel formulation for the joint periodic demand estimation and tactical planning problem. \ef{For the first step we rely on state-of-the-art forecasting models, while the second step constitutes the core of our methodological contribution.} In the following section\greta{,} we describe assumptions and their implications on the time series forecasting problem. In Section~\ref{sec:meth_perdemand}, we delineate the mathematical programming formulation.

\subsection{Time Series Forecasting} \label{sec:meth_forecasting}

\greta{In this section we focus on the time series forecasting problem. Our objective is to predict, at period $t$, $T$ point estimates $\hat{\mathbf{y}}_{t+1}, \dots, \hat{\mathbf{y}}_{t+T}$. We consider a medium-term forecasting horizon (two to three months), hence we require a tactical precision level.}
\ef{However,} as we highlight in the previous section, historical data capture operational flows which can be constrained by the supply\ef{.}
\greta{Hence, for a given $t$, the time series of operational data records $\{y_{1k}^t, \ldots,y_{dk}^t,\ldots, y_{Dk}^t\}$ may be subject to truncation or censoring. Moreover, the observations can be right or left censored. Consider, on the one hand, a service that has reached its capacity and the operations are forced to leave demand behind. This would result in a right censored observation. On the other hand, consider the services used to transport the demand left behind, those observations would be left censored. The setting is therefore complex\gl{,} and we do not assume any knowledge on the process leading to censoring or truncation \citep[this prevents the use of some existing statistical methods, such as the one proposed by][]{FieldsEtAl21Trunc}. We note, however, that detailed data on operational decisions could provide at least partial information on that process. Instead, we adopt a data aggregation approach to deal with censoring/truncation. If two weak assumptions are satisfied, we can use \emph{aggregate} historical \emph{uncensored} data for time series forecasting.}

Recall from the problem description that demand which cannot be satisfied by the planned capacity is outsourced. This is typically the case for carriers as unsatisfied demand would otherwise accumulate over time periods. Taking into account the outsourcing, demand is hence assumed satisfied in each time period $t$. This leads us to the following assumption on the historical data.
\begin{corollary}
Historical data are uncensored when aggregated over time periods. That is, $\mathbf{y}_{s}=\sum_{d=1}^{D_s} \mathbf{y}_d^s, s=t-1,t-2,\ldots,t-H$ are uncensored.
\label{hyp:assumption_censoring}
\end{corollary}
\ef{If this assumption is satisfied, we do not need statistical uncensoring methods but} aggregation results in fewer data points to learn from.
\greta{In this context we describe two important trade-offs that we consider. First, the one between data quantity and variance. Using operational (e.g., daily) as opposed to aggregate data results in more observations. However, operational observations typically have higher variance. Second, the trade-off between the prediction granularity/precision and the length of the prediction horizon. We focus on tactical planning and hence forecast over a horizon that far exceeds that of operational planning. Forecasting over a tactical horizon (e.g., two to three months) but at an operational precision (e.g., daily) may lead to unnecessarily large variance in prediction errors. We now introduce an assumption related to this second trade-off.}

The tactical planning problem formulation is based on a space-time graph. The departure and arrival times of a commodity $k$ are hence implicitly given by $o_k$ and $d_k$. We assume that the arrival and departure times are endogenous decisions, stated in other words\greta{,} in the following assumption.
\begin{corollary}
Predicted demand per time period for each commodity, $\hat{\mathbf{y}}_t,~t=1,\ldots,T$, are sufficiently precise for tactical planning.
\label{hyp:hyp2}
\end{corollary}
This is a weak assumption considering that, if exogenous predictions of $\hat{\mathbf{y}}_d^t,~d=1,\ldots,D_t$ are required for each $t=1,\ldots,T$, it is possible to define a model (different from the time series one) that projects $\hat{\mathbf{y}}_t$ down to that level. \greta{We note that the latter would require an uncensoring method.}

\subsection{A Multilevel Formulation} 
\label{sec:meth_perdemand}

We define the feasible set of periodic demand vectors
\begin{equation}
    \mathcal{Y} = \{ \hat{\mathbf{y}}^{\text{p}} = h_i(\hat{\mathbf{y}}_1, \ldots \hat{\mathbf{y}}_T), i=1,\ldots,I \} 
    \label{eq:periodic_set}
\end{equation}
by a finite set of mappings $h_i$, $~i=1,\ldots,I$ ~\eqref{eq:h_mapping}. We propose the following multilevel formulation \textbf{PDE} for the periodic demand estimation problem.  

\begin{small}
\begin{flalign}
    \textbf{PDE} \quad \min_{\hat{\mathbf{y}}^{\text{p}}} & \quad C^{\text{PDE}} = \sum_{t=1}^T \left[ \sum_{p \in \mathcal{P}} C_p^{\text{design}} z_p  +  \sum_{k \in \mathcal{K}} \sum_{p \in \mathcal{P}_k \setminus \mathcal{P}_k^{\text{out}}} C_p^{\text{flow}} x_{tpk} +  \sum_{k \in \mathcal{K}} \sum_{p \in \mathcal{P}_k^{\text{out}}}  C_p^{\text{out}} x_{tpk} \right] \label{eq:genericP_obj} \\
    & \nonumber \\
    \text{s.t.} & \hspace{0.3cm} \hat{\mathbf{y}}^{\text{p}} \in \mathcal{Y} \\
    \textbf{MCND} & \hspace{0.3cm} \min_{z, x} \sum_{p \in \mathcal{P}} C_p^{\text{design}} z_p   +  \sum_{k \in \mathcal{K}} \sum_{p \in \mathcal{P}_k \setminus \mathcal{P}_k^{\text{out}}} C_p^{\text{flow}} x_{pk} +  \sum_{k \in \mathcal{K}} \sum_{p \in \mathcal{P}_k^{\text{out}}}  C_p^{\text{out}} x_{pk} \label{eq:genericP_l1_obj} \\
    & \hspace{0.3cm} \text{ s.t. } \sum_{p \in \mathcal{P}_k} x_{pk} = \hat{y}_k^{\text{p}}, \hspace{2.3cm} k \in \mathcal{K}, \label{eq:genericP_l1_demand}\\ 
    & \hspace{0.8cm} \sum_{k \in \mathcal{K}_p} x_{pk} \leq u_p z_p, \hspace{2cm} p \in \mathcal{P}, \label{eq:genericP_l1_cap}\\ 
    & \hspace{0.8cm} x_{pk} \geq 0 , \hspace{3.2cm} k \in \mathcal{K}, p \in \mathcal{P}_k,  \label{eq:genericP_l1_flowPos}\\ 
    & \hspace{0.8cm} z_p \in \{0,1\}, \hspace{2.8cm}  p \in \mathcal{P},  \label{eq:genericP_l1_design}\\
    \textbf{wMCND} & \hspace{0.8cm} \min_{x_1,\ldots,x_T} \sum_{t=1}^T \left[ \sum_{k \in \mathcal{K}} \sum_{p \in \mathcal{P}_k \setminus \mathcal{P}_k^{\text{out}}} C_p^{\text{flow}} x_{tpk} +  \sum_{k \in \mathcal{K}} \sum_{p \in \mathcal{P}_k^{\text{out}}}  C_p^{\text{out}} x_{tpk} \right] \label{eq:genericP_l2_obj} \\
    & \hspace{0.8cm} \text{ s.t. } \sum_{p \in \mathcal{P}_k} x_{tpk} = \hat{y}_{tk}, \hspace{1.6cm} t=1,\ldots,T, k \in \mathcal{K}, \label{eq:genericP_l2_demand}\\ 
    & \hspace{1.4cm} \sum_{k \in \mathcal{K}_p} x_{tpk} \leq u_p z_p, \hspace{1.3cm} t=1,\ldots,T, p \in \mathcal{P}, \label{eq:genericP_l2_cap}\\ 
    & \hspace{1.4cm} x_{tpk} \geq 0 , \hspace{2.55cm} t=1,\ldots,T, k \in \mathcal{K}, p \in \mathcal{P}_k. \label{eq:genericP_l2_flowPos}
\end{flalign}
\end{small}

The upper level selects $\hat{\mathbf{y}}^{\text{p}}$ that minimizes the total fixed and variable costs over the whole tactical planning horizon.   The objective function~\eqref{eq:genericP_obj} hence depends on the design and flow variables from the lower levels \textbf{MCND} and \textbf{wMCND}, respectively. \efsec{We note that fixed costs incur in each time period as they correspond to cost associated with running services. The objective function could, however, be easily modified to the case where fixed costs incur once over the time horizon.}
This multilevel formulation allows to obtain important information for the carrier: the periodic demand $\hat{\mathbf{y}}^{\text{p}}$ and the plan repeated in each period.
%given by the solutions $z$ and $x$ of the second level \textbf{MCND}.
\textbf{MCND} and \textbf{wMCND} form two sequential problems. Moreover, they have a different temporal interpretation. While \textbf{MCND} is a single-period problem, \textbf{wMCND} is a multi-period  problem (there are $T$ periods). In Figure~\ref{fig:planning_processes}, where the length of the planning horizon goes from medium term (left-hand side) to short term (right-hand side), \textbf{MCND} corresponds to the \efsec{SND} problem, while \textbf{wMCND} serves as a proxy for the operational planning problem (framed with a blue, dotted, line \efsec{in the figure}). 

In \textbf{wMCND} we introduce the flow variables $x_{tpk}$ for commodity $k$ on path $p$ in period $t$ and determine flows for each period minimizing variable cost~\eqref{eq:genericP_l2_obj} for a fixed design solution $z$ given by \textbf{MCND}. Constraints~\eqref{eq:genericP_l2_demand} ensure that the demand is satisfied for each commodity in each period. The set of paths $\mathcal{P}_k$ includes outsourcing paths $\mathcal{P}_k^{\text{out}}$ for a commodity $k$, so constraints \eqref{eq:genericP_l1_demand} and \eqref{eq:genericP_l2_demand} can always be satisfied. Constraints~\eqref{eq:genericP_l2_cap} enforce flows in each period to be only on selected paths and smaller than the capacity of the path. We draw the attention to the time series forecasts that occur in \textbf{wMCND} while the periodic demand estimates occur in \textbf{MCND}.

The decision variables of \textbf{wMCND} do not occur in the objective function~\eqref{eq:genericP_l1_obj} of \textbf{MCND}. Thus, for $(z^*, x^*)$ an optimal solution of \textbf{MCND}, if $z^*$ is feasible for~\textbf{wMCND}, then $z^*$ is an optimal solution to \text{\textbf{MCND}-\textbf{wMCND}}. We can therefore make the following claim.

\begin{theorem} \label{claim:seq}
If $z^*$ is feasible for \textbf{wMCND}, then \text{\textbf{MCND}-\textbf{wMCND}} can be solved sequentially to optimality for a fixed $\hat{\mathbf{y}}^{\text{p}}$.
\end{theorem}

\efsec{In other words, for a given $\hat{\mathbf{y}}^{\text{p}}$, we first solve \textbf{MCND} to obtain $z^*$ (and $x^*$). We then solve \textbf{wMCND} with design variables fixed to $z^*$, hence obtaining $x_1^*,x_2^*,\ldots,x_T^*$. We note that the feasibility assumption is weak, as outsourcing, albeit costly, is generally possible in practice. When the set of paths contains such outsourcing paths, all demand can be satisfied. If the feasibility assumption does not hold, it has to be solved as a bilevel problem which is considerably more involved \citep[see, e.g.,][for a survey]{ColsEtAl05}.}

In this work, we consider four mappings \efsec{in $\mathcal{Y}$},
\begin{equation}
    \mathbf{y}_{\text{max}}^{\text{p}} = h_1(\mathbf{y}_1, \ldots, \mathbf{y}_T)= \max_{t=1,\ldots,T} \mathbf{y}_t,
    \label{eq:max_periodic}
\end{equation}
\begin{equation}
    \mathbf{y}_{\text{mean}}^{\text{p}} =  h_2(\mathbf{y}_1, \ldots, \mathbf{y}_T) =  \frac{1}{T} \sum_{t=1}^T \mathbf{y}_t,
    \label{eq:mean_periodic}
\end{equation}
\begin{equation}
    \mathbf{y}_{\text{q2}}^{\text{p}} =  h_3(\mathbf{y}_1, \ldots, \mathbf{y}_T) =  Q_2(\mathbf{y}_{t}, t = 1 \ldots, T),
    \label{eq:q2_periodic}
\end{equation}
\begin{equation}
    \mathbf{y}_{\text{q3}}^{\text{p}} =  h_4(\mathbf{y}_1, \ldots, \mathbf{y}_T) =  Q_3(\mathbf{y}_{t}, t = 1 \ldots, T),
    \label{eq:q3_periodic}
\end{equation}
\greta{representing} the maximum, mean, second quartile $Q_2$ and third quartile $Q_3$, respectively. The corresponding estimates from the forecasts are denoted
$\hat{\mathbf{y}}_{\text{max}}^{\text{p}}$, $\hat{\mathbf{y}}_{\text{mean}}^{\text{p}}$, $\hat{\mathbf{y}}_{\text{q2}}^{\text{p}}$ and $\hat{\mathbf{y}}_{\text{q3}}^{\text{p}}$. Given that we consider a discrete $\mathcal{Y}$ of small cardinality, we find the solution to \textbf{PDE} by solving \textbf{MCND}-\textbf{wMCND} for each $\mathbf{y}^{\text{p}}\in\mathcal{Y}$. \efsec{The mappings may be seen as a restriction on the full feasible space (any feasible periodic demand vector). For a given computing time budget, there is hence a trade-off between the time it takes to solve \textbf{MCND}-\textbf{wMCND} and the number of considered mappings. Higher quality solutions may be found by increasing the number of considered mappings. The choice of mappings can therefore be adapted to each application. As we report in Section~\ref{section:results}, for our application, we find solutions that lead to substantial cost reductions using the aforementioned mappings.}

In the following section, we describe a specific instance of~\eqref{eq:genericP_obj}-\eqref{eq:genericP_l2_flowPos} in the context of tactical planning for intermodal rail transportation.

\section{Application}
\label{section:block_plan}

\ef{In order to apply the proposed methodology, we need an application for which we have access to historical data on demand as well as a formulation of a \efsec{SND} problem. We consider intermodal (container) transportation by rail at our industrial partner, CN. As we further detail in Section~\ref{section:results}, we have access to six years of data. Moreover, \cite{Morganti2019} describe a specific instance of \textbf{MCND} designed to solve the so-called block planning problem. However, they solely focus on the optimization problem and take daily demand values as fixed and known. In other words, they ignore the need for demand forecasts and hence the problem we focus on in this paper. We therefore use their formulation as an application for our methodology. In this section we start by briefly describing CN's operations and the formulation proposed in \cite{Morganti2019}. Then, in Section~\ref{sec:weeklyDemandBP}, we slightly adapt the formulation such that it takes weekly, as opposed to daily, demand as input. Finally, in Section~\ref{sec:PDE_BP} we introduce the application-specific \textbf{PDE} formulation.}

\begin{figure}[htbp]
\centering
  \includegraphics[width=0.6\linewidth]{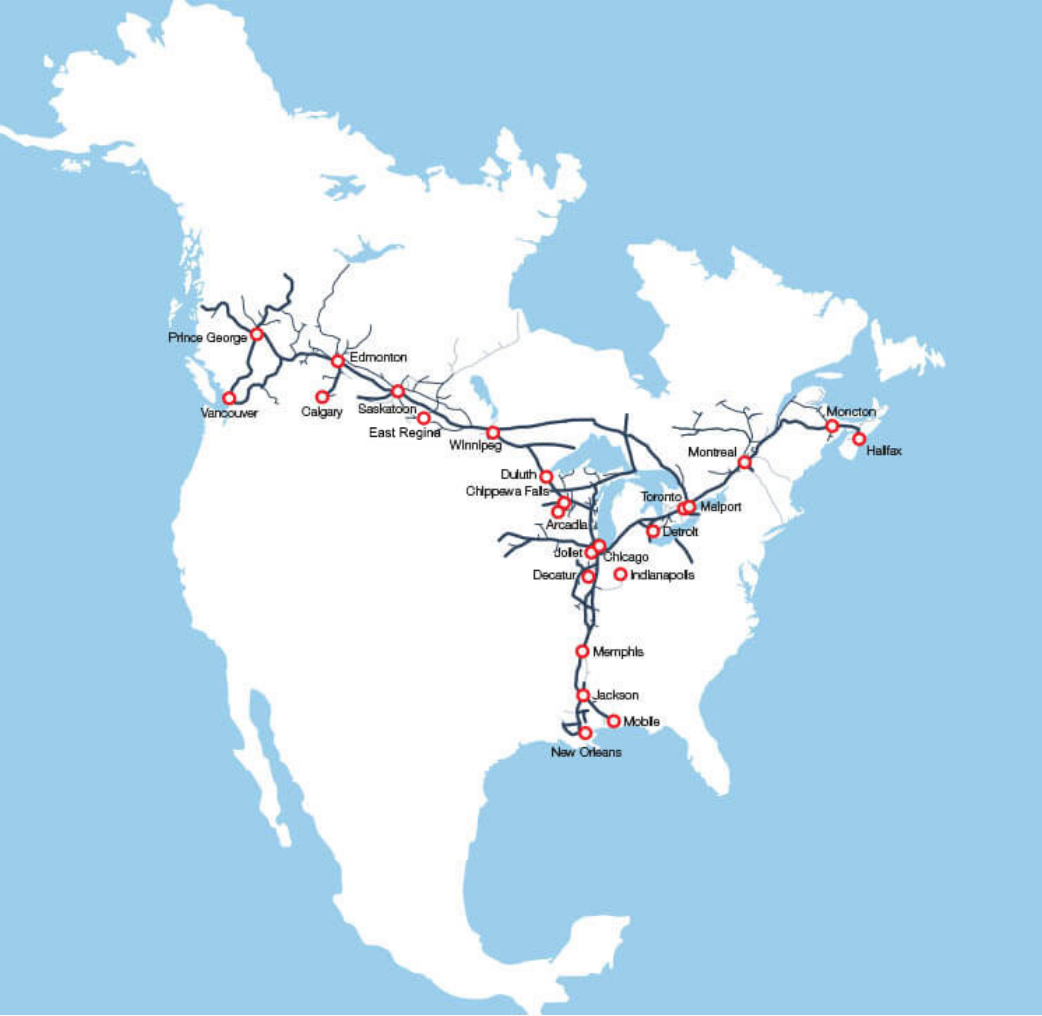}
  \caption{Intermodal Network of the Canadian National Railway Company. Source: {www.cn.ca}}
  \label{fig:intermodal_net}
\end{figure}

The intermodal network of CN is composed of 24 main intermodal terminals and 133 origin-destination (OD) pairs. Figure~\ref{fig:intermodal_net} depicts a map of the network. The rail tracks extend from \ef{Eastern} to \ef{Western} Canada and from Canada to \ef{Southern} United States. The railroad carries a variety of container types (20, 40, 45, 48 and 53-feet long), yet for tactical planning purposes they can be aggregated into either 40-feet or 53-feet containers. Indeed, 20-feet containers can be considered as half-40-feet containers. Other sizes (45, 48 and 53-feet) occupy the same locations on railcars \citep[so-called slots,][]{MantEtAl18} as 53-feet containers. A commodity is defined by an origin, a destination and container \efsec{type}, and we consider a total of $K = 170$ commodities. Two commodities can hence have the same OD pair but differ by container \efsec{type}. The tactical period is a week and a tactical horizon lasts $T = 10$ weeks. The train schedule is repeated each week and CN operates so that demand over a week is satisfied.  Assumption~\ref{hyp:assumption_censoring} therefore holds. More precisely, in case of insufficient capacity on intermodal trains, they use general cargo trains or additional ad hoc intermodal trains to satisfy demand.

\cite{Morganti2019} introduce a path-based Block Planning formulation (\textbf{BP}) \ef{which is a specific instance of \textbf{MCND}. A block refers to a consolidation of railcars. In this context, \greta{a block is} a set of railcars flowing as a single unit between a given OD pair and where containers loaded on the railcars have the same OD. The formulation is} defined \ef{based on} a space-time graph generated \greta{from} a schedule of intermodal trains \greta{and CN's physical network}. The graph contains 28,854 arcs and 15,269 nodes. A block is a path in this graph and the set is denoted $\mathcal{B}$ with $|\mathcal{B}|=2,208$. We keep this notation to be consistent with \cite{Morganti2019} but note that the set $\mathcal{B}$ corresponds to the set of paths $\mathcal{P}$ in \textbf{MCND}. The set $\mathcal{B}$ contains a subset of \textit{artificial blocks} $\mathcal{B}^{\text{artif}}$ whose role is to transport demand exceeding capacity. They hence correspond to the outsourcing paths. They are built without design cost, i.e., $C_b^{\text{design}} = 0, \forall b \in \mathcal{B}^{\text{artif}}$. Similarly to \textbf{MCND}, $\mathcal{B}_k$ and $\mathcal{B}_k^{\text{artif}}$ denote respectively the set of blocks and the set of artificial blocks for commodity $k$,  and $\mathcal{K}_b$ the set of commodities that can use $b$.

Below we briefly describe \textbf{BP} and refer to \cite{Morganti2019} for more details. There are three categories of decision variables. First, the design variables $z_b, b \in \mathcal{B}$ where $z_b$ equals one if block $b$ is built, and zero otherwise. Second, integer flow variables $x_{bk},k\in\mathcal{K},b\in\mathcal{B}_k$ that equal the number of containers for commodity $k$ transported on block $b$. Third, auxiliary variables for the number of 40-feet $v_b^{40}$ and 53-feet $v_b^{53}$ double-stack platforms to carry the containers assigned to block $b\in\mathcal{B}$. 
\begin{flalign}
    \textbf{BP} & \hspace{0.2cm} \min_{x, z} \sum_{b \in \mathcal{B} \setminus \mathcal{B}^{\text{artif}} } C_b^{\text{design}} z_b +  \sum_{k \in \mathcal{K}} \sum_{b \in \mathcal{B}_k \setminus \mathcal{B}_k^{\text{artif}} } C_{bk}^{\text{flow}} x_{bk} + \sum_{k \in \mathcal{K}} \sum_{b \in \mathcal{B}_k^{\text{artif}} } C_{bk}^{\text{out}} x_{bk}  \label{eq:objFun} \\ 
    & \hspace{0.2cm} \text{ s.t. } \sum_{b \in \mathcal{B}_k} x_{bk} = y_k^{\text{p}},  \hspace{5.45cm}  k \in \mathcal{K}, \label{eq:demand}\\
    & \hspace{0.8cm} x_{bk} \leq y_k^{\text{p}} z_b, \hspace{5.95cm}  k \in \mathcal K, b \in \mathcal{B}_k, \label{eq:flow}\\
    & \hspace{0.8cm} v_{b}^{53} =  \max \left [ 0, \left \lceil \frac{1}{2} \left ( \sum_{k \in \mathcal{K}_b, \tau_k = 53} x_{bk} - \sum_{k \in \mathcal{K}_b, \tau_k=40 } x_{bk} \right ) \right \rceil \right ],  \nonumber \\
    & \hspace{8.4cm}  b \in \mathcal{B}, \label{eq:cap1}\\
    & \hspace{0.8cm} v_{b}^{40} = \left \lceil \frac{1}{2} \left ( \sum_{k \in \mathcal{K}_b} x_{bk}  \right ) \right \rceil - v_{b}^{53}, \hspace{3cm}  b \in \mathcal{B}, \label{eq:cap2} \\
    & \hspace{0.8cm} \sum_{b \in \mathcal{B}_a} \left( L^{40} v_b^{40} + L^{53} v_b^{53} \right) \leq u_a, \hspace{3.02cm}  a \in \mathcal{A}^{TM},  \label{eq:cap3} \\
    & \hspace{0.8cm} z_{b} \in \{0,1\}, \hspace{5.85cm}  b \in \mathcal{B},  \\
    & \hspace{0.8cm} v_b^{40}, v_b^{53} \in \mathbb{N}, \hspace{5.65cm}  b \in \mathcal{B}, \\
    & \hspace{0.8cm} x_{bk} \in \mathbb{N}, \hspace{6.25cm}  k \in \mathcal K, b \in \mathcal{B}_k. \label{eq:bp_final}
\end{flalign}
The objective function~\eqref{eq:objFun} minimizes fixed and variable costs as well as a variable cost associated with outsourced demand (flow on artificial blocks). Constraints~\eqref{eq:demand} ensure that the demand is satisfied by either the network capacity or outsourcing. Constraints~\eqref{eq:flow} enforce flows to be on selected blocks only. Constraints~\eqref{eq:cap1} and~\eqref{eq:cap2} fix the number of platforms required to transport the demand. These constraints take into account how containers of different sizes can be double stacked. Since 40-feet platforms use less train capacity than 53-feet platforms, they are used whenever there are less 53-feet containers than 40-feet ones (40-feet container stacked in the bottom position and 53-feet container on top), and 53-feet platforms are used otherwise. Constraints~\eqref{eq:cap3} ensure that the train capacity, expressed in number of feet, is not exceeded. The platform lengths are denoted $L^{40}$ and $L^{53}$, respectively. The train capacity, $u_a, a\in \mathcal{A}^{TM}$, is defined for the set of arcs in the space-time graph that represent moving trains, $\mathcal{A}^{TM}$. We denote by $\mathcal{B}_a$ the set of blocks that use train moving arc $a \in \mathcal{A}^{TM}$. 

Finally, we give an order of magnitude of the size of the formulation \gl{\textbf{BP}}. In the case of the instances we solve in this paper, there are over 386,000 variables and some 18,000 constraints.

\subsection{Block Generation For Weekly Demand Inputs} \label{sec:weeklyDemandBP}

\begin{figure}[htbp]  
  \begin{subfigure}[b]{0.5\linewidth}
    \begin{tikzpicture}
    \node[shape=circle,draw=black, fill = gray!100, scale = 0.8] (A) at (1,1) {DIN};
    \node[shape=circle,draw=black, draw = white, fill = white] (B) at (1,2.3) {};
    \node[shape=circle,draw=black, draw = white, fill = white] (C) at (5, 2.3) {};
    \node[shape=circle, draw=black, fill = gray!100, scale = 0.8] (D) at (5, 1) {DIN'};
    \node[shape=circle,draw=black, draw = white, fill = white] (E) at (-0.1, -0.5) {};
    \node[shape=circle,draw=black, draw = white, fill = white] (F) at (3.5, -0.5) {};
    \path [densely dashed, ->] (A) edge node[left] {} (B);
    \path [->](A) edge node[below] {\textit{ContainersWaiting}} (D);
    \path [densely dashed, ->](D) edge node[left] {} (C);
    \path [->] (E) edge[ultra thick] node[right] {\begin{tabular}{c}Daily \\Demand In \\\end{tabular}} (A);
    \path [->] (F) edge[ultra thick] node[right] {\begin{tabular}{c}Daily \\Demand In \\\end{tabular}} (D);
\end{tikzpicture}
    \caption{\cite{Morganti2019}}
    \label{fig:container1}  
  \end{subfigure}
\begin{subfigure}[b]{0.5\linewidth}
\begin{tikzpicture}
    \node[shape=circle,draw=black, fill = gray!100, scale = 0.8] (A) at (1,1) {DIN};
    \node[shape=circle,draw=black, draw=black, draw = white] (B) at (1,2.3) {};
    \node[shape=circle,draw=black, draw=black, draw = white] (C) at (5, 2.3) {};
    \node[shape=circle, draw=black, fill = gray!100, scale = 0.8] (D) at (5, 1) {DIN'};
    \node[shape = circle, draw = black, fill = green!20, scale = 0.8] (G) at (1, -1){$n_{\theta}^{\text{WIN}}$};
    \node[shape=circle,draw=black, draw = white, fill = white] (E) at (-0.1, -2.5) {};
    
    \path [densely dashed, ->] (A) edge node[left] {} (B);
    \path [densely dashed, ->](D) edge node[left] {} (C);
    \path [->] (G) edge node[right] {$a \in \mathcal{A}^{\text{WIN}}$} (A);
    \path [->] (G) edge node[right] {$a' \in \mathcal{A}^{\text{WIN}}$} (D);
    \path [->] (E) edge[ultra thick] node[right] {\begin{tabular}{c}Weekly  \\Demand In \\\end{tabular}} (G);
\end{tikzpicture}
\caption{Our model} 
\label{fig:container2}  
\end{subfigure}
\caption{Illustration of the difference between \cite{Morganti2019} and our model}
\label{fig:container_layer}
\end{figure}
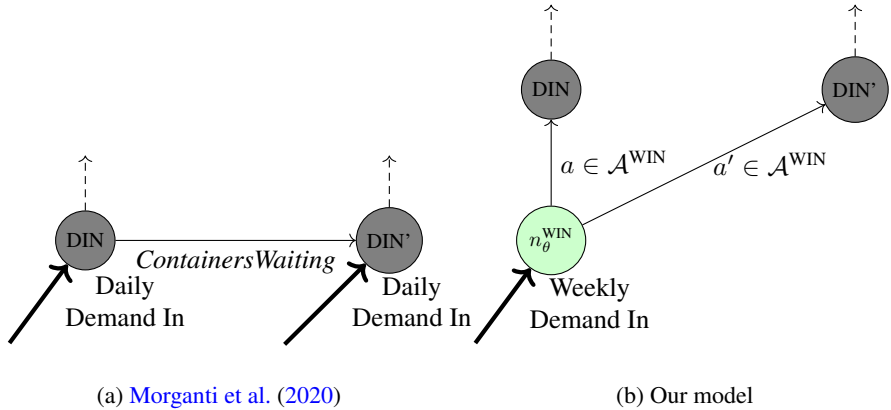

While the tactical plan is computed for a weekly schedule, \cite{Morganti2019} assume that the time at which the demand arrives to the system within the week is given exogenously. We provide an illustrative example in Figure~\ref{fig:container1}. Demand enters the network via a node noted DIN which is associated to one admissible train departure node. Containers are either assigned to a block, or wait at the terminal, represented by flow on arcs called \textit{ContainersWaiting}. 

Under Assumption~\ref{hyp:hyp2}, we propose a slightly different block generation so that the model optimally distributes the weekly demand over the train departures. 
For this purpose, we introduce a new set of nodes $\mathcal{N}^{\text{WIN}}$ such that there is one node WIN $n_{\theta}^{\text{WIN}} \in \mathcal{N}^{\text{WIN}}$ per terminal $\theta \in \Theta$, where $\Theta$ is the set of terminals in the network. We also introduce a new set of arcs 
$\mathcal{A}^{\text{WIN}} = \{ (n_{\theta}^{\text{WIN}}, j) \mid \theta \in \Theta, j \in \mathcal{N}^{\text{DIN}}_{\theta}\}$, where $\mathcal{N}^{\text{DIN}}_{\theta}$ is the set of DIN nodes for terminal $\theta$.

We illustrate this change compared to \cite{Morganti2019} in Figure~\ref{fig:container2}. For each terminal, the node WIN receives the weekly demand input and splits the commodity flow to the different DIN nodes on the arcs $\mathcal{A}^{\text{WIN}}$ at no cost. Instead of arriving at different points in time, demand now arrives in one source node and the model selects the optimal distribution over the week.

\subsection{Periodic Demand Estimation Problem} \label{sec:PDE_BP}

\ef{In this section w}e present the \ef{application-specific \textbf{PDE}} formulation.  
The \textbf{MCND} formulation is replaced by \textbf{BP}\ef{. The formulation has two differences compared to \cite{Morganti2019}. First, we change the block generation as described in Section~\ref{sec:weeklyDemandBP}. Second, and more importantly, demand is represented by first level decision variables $\hat{\mathbf{y}}^{\text{p}}$ as opposed to parameters representing ground truth demand values. Moreover,} we introduce a weekly \ef{flow} formulation, \textbf{wBP}. 
\greta{For this third level} we introduce flow variables and auxiliary platform variables for each week $t$, $x_{tbk}, v_{tb}^{40}, v_{tb}^{53}, t \in \mathcal{T},  k \in \mathcal{K},  b \in \mathcal{B}$.

\begin{footnotesize}
\begin{flalign}
    \textbf{PDE BP} \min_{\hat{\mathbf{y}}^{\text{p}}} & \hspace{0.02cm} C^{\text{PDE}} = \sum_{t=1}^T \left[ \sum_{b \in \mathcal{B} \setminus \mathcal{B}^{\text{artif}} } C_b^{\text{design}} z_b + \sum_{k \in \mathcal{K}} \sum_{b \in \mathcal{B}_k \setminus \mathcal{B}_k^{\text{artif}} } C_{tbk}^{\text{flow}} x_{tbk} + \sum_{k \in \mathcal{K}} \sum_{b \in \mathcal{B}_k^{\text{artif}} } C_{tbk}^{\text{out}} x_{tbk} \right] \label{eq:objP1} \\
    & \nonumber \\
    \text{s.t.} & \hspace{0.05cm} \hat{\mathbf{y}}^{\text{p}} \in \mathcal{Y}, \label{eq:pde_bp_2}\\
    \textbf{BP} & \hspace{0.05cm} \min_{x, z} \sum_{b \in \mathcal{B} \setminus \mathcal{B}^{\text{artif}} } C_b^{\text{design}} z_b + \sum_{k \in \mathcal{K}} \sum_{b \in \mathcal{B}_k \setminus \mathcal{B}_k^{\text{artif}} } C_{bk}^{\text{flow}} x_{bk} + \sum_{k \in \mathcal{K}} \sum_{b \in \mathcal{B}_k^{\text{artif}} } C_{bk}^{\text{out}} x_{bk} \label{eq:objFun_BP} \\ 
    & \hspace{0.05cm} \text{ s.t. } \sum_{b \in \mathcal{B}_k} x_{bk} = \hat{y}_k^{\text{p}},  \hspace{5.1cm}  k \in \mathcal{K}, \label{eq:periodic_bp}\\
    & \hspace{0.5cm} x_{bk} \leq \hat{y}_k^{\text{p}} z_b, \hspace{5.6cm}  k \in \mathcal{K},  b \in \mathcal{B}_k,  \\
    & \hspace{0.5cm} v_{b}^{53} =  \max \left [ 0, \left \lceil \frac{1}{2} \left ( \sum_{k \in \mathcal{K}_b, \tau_k = 53} x_{bk} - \sum_{k \in \mathcal{K}_b, \tau_k=40 } x_{bk} \right ) \right \rceil \right ],    \nonumber \\
    & \hspace{7.6cm} b \in \mathcal{B}, \label{eq:cap1_bis}\\
    & \hspace{0.5cm} v_{b}^{40} = \left \lceil \frac{1}{2} \left ( \sum_{k \in \mathcal{K}_b} x_{bk}  \right ) \right \rceil - v_{b}^{53}, \hspace{2.95cm}  b \in \mathcal{B}, \label{eq:cap2_bis} \\
    & \hspace{0.5cm} \sum_{b \in \mathcal{B}_a} \left( L^{40} v_b^{40} + L^{53} v_b^{53} \right) \leq u_a, \hspace{3cm}  a \in \mathcal{A}^{TM},  \label{eq:cap3_bis} \\
    & \hspace{0.5cm} z_{b} \in \{0,1\}, \hspace{5.6cm}  b \in \mathcal{B},  \\
    & \hspace{0.5cm} v_b^{40}, v_b^{53} \in \mathbb{N}, \hspace{5.4cm}  b \in \mathcal{B}, \\
    & \hspace{0.5cm} x_{bk}  \in \mathbb{N}, \hspace{5.95cm}  k \in \mathcal{K}, b \in \mathcal{B}_k, \label{eq:pde_bp_end1}\\
    \textbf{wBP} & \hspace{0.6cm} \min_{x_1, \ldots, x_T} \sum_{t = 1}^T \sum_{k \in \mathcal{K}} \left[ \sum_{b \in \mathcal{B}_k \setminus \mathcal{B}_k^{\text{artif}} } C_{tbk}^{\text{flow}} x_{tbk} + \sum_{b \in \mathcal{B}_k^{\text{artif}} } C_{tbk}^{\text{out}} x_{tbk} \right] \label{eq:pde_wbp_1}  \\ 
    & \hspace{0.6cm} \text{ s.t. } \sum_{b \in \mathcal{B}_k} x_{tbk} = \hat{y}_{tk},  \hspace{4.3cm}  t \in \mathcal{T},  k \in \mathcal{K}, \label{eq:pde_wbp_2}\\
    & \hspace{1.2cm} x_{tbk} \leq \hat{y}_{tk} z_b, \hspace{4.6cm}  t \in \mathcal{T},  k \in \mathcal{K},  b \in \mathcal{B}_k,  \\
    & \hspace{1.2cm} v_{tb}^{53} =  \max \left [ 0, \left \lceil \frac{1}{2} \left ( \sum_{k \in \mathcal{K}_b, \tau_k = 53} x_{tbk} - \sum_{k \in \mathcal{K}_b, \tau_k=40 } x_{tbk} \right ) \right \rceil \right ], \nonumber\\
    & \hspace{7.55cm} t \in \mathcal{T},  b \in \mathcal{B}, \label{eq:bp_platf53_2} \\
    & \hspace{1.2cm} v_{tb}^{40} = \left \lceil \frac{1}{2} \left ( \sum_{k \in \mathcal{K}_b} x_{tbk}  \right ) \right \rceil - v_{tb}^{53}, \hspace{2.12cm}  t \in \mathcal{T},   b \in \mathcal{B},  \label{eq:bp_platf40_2}\\
    & \hspace{1.2cm} \sum_{b \in \mathcal{B}_a} \left( L^{40} v_{tb}^{40} + L^{53} v_{tb}^{53} \right) \leq u_a, \hspace{2.25cm}  t \in \mathcal{T},  a \in \mathcal{A}^{TM},   \\
    & \hspace{1.2cm} v_{tb}^{40}, v_{tb}^{53} \in \mathbb{N}, \hspace{4.58cm}  t \in \mathcal{T}, b \in \mathcal{B} \label{eq:last_1_P}, \\
    & \hspace{1.2cm} x_{tbk} \in \mathbb{N}, \hspace{5cm}  t \in \mathcal{T},  k \in \mathcal{K},  b \in \mathcal{B}_k. \label{eq:last_P}
\end{flalign}
\end{footnotesize}
The objective function~\eqref{eq:objP1} has the same structure as~\eqref{eq:genericP_obj}, with design costs, flow costs and outsourcing costs.  
 We note that we define \textbf{BP}~\eqref{eq:objFun_BP}-\eqref{eq:pde_bp_end1} using periodic demand $\hat{\mathbf{y}}^{\text{p}}$, while we define \textbf{wBP}~\eqref{eq:pde_wbp_1}-\eqref{eq:last_P} for fixed design variables and weekly demand $\hat{\mathbf{y}}_t$.
Furthermore, there is no fixed cost associated with $b \in \mathcal{B}^{\text{artif}}$. Therefore, a solution for \textbf{BP} is always feasible for \textbf{wBP}. Using Claim~\ref{claim:seq}, we can solve \textbf{BP}-\textbf{wBP} sequentially.

\section{Computational Results}
\label{section:results}

Our dataset contains the observed daily container shipments of all types of containers on each \efsec{OD} pair of the rail network, collected over 6 years, from December 2013 to November 2019. The tactical plan is weekly. To ensure the observed demand is not constrained by the supply, we do a weekly aggregation of 2,226 observations of daily demand for each commodity. This results in 318 observations per commodity. Hence, $y_{tk}$ refers to the number of containers of type $\tau_k$ to be carried from origin $o_k$ to destination $d_k$ during week $t$. 

In this section, we first provide a descriptive analysis of the data. We report the results of the time series forecasting models in Section~\ref{subsec:fcst_model_results}. Finally, in Section~\ref{subsec:bp_results} we report results for the periodic demand estimation problem. Note that, for confidentiality reasons, we only report relative numbers in all results. 

\subsection{Descriptive Analysis}
\label{subsec:data}

The large size of the transportation network and the large number of commodities suggest \greta{the presence of spatio-temporal correlation}. We provide here an analysis of the different types of correlations we identified: between commodities and between commodities and weather. 

\subsubsection{Correlation Between Commodities} \label{sec:correlationComm}

\begin{figure}[htbp]
    \begin{subfigure}[b]{0.5\linewidth}
        \includegraphics[width=\linewidth]{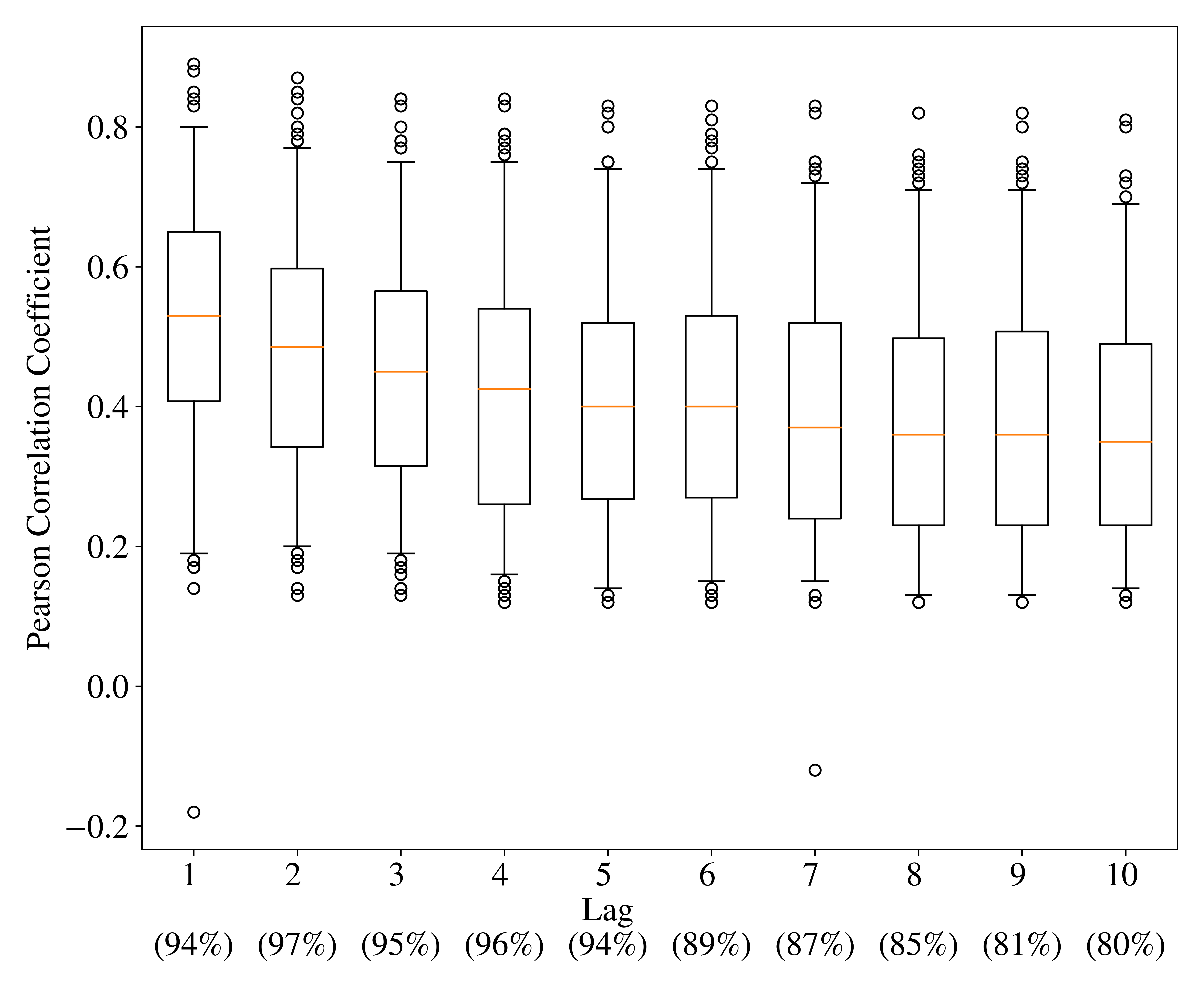}
        \caption{\tabular[t]{@{}l@{}} Distribution of interweek correlations \\ over commodities. \endtabular}
        \label{fig:interweek_correlations}  
    \end{subfigure}
    \begin{subfigure}[b]{0.5\linewidth}
        \includegraphics[width=\linewidth]{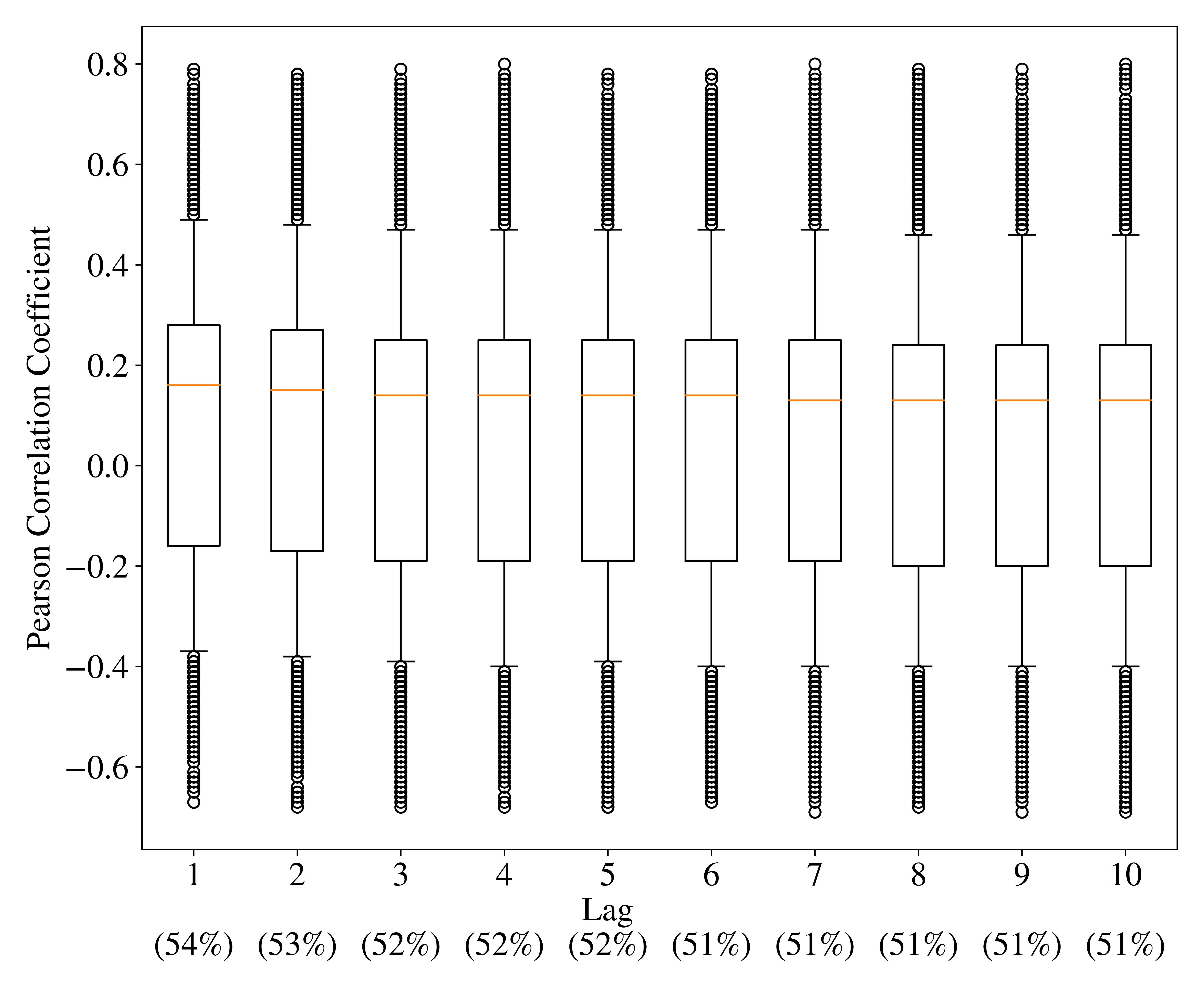}
        \caption{\tabular[t]{@{}l@{}} Distribution of intercommodity correlations  \\ over commodities. \endtabular}
        \label{fig:intertype_commodities}
    \end{subfigure}
\caption{Distribution of the Pearson correlation coefficient for the two different types of correlations. We consider only the coefficients which met the 95\% confidence level threshold and we indicate their proportion in parentheses.}
\label{fig:correlations}
\end{figure}

We start by analyzing interweek correlation by computing, for each commodity, estimates of the Pearson correlation coefficient between weekly shipments over successive weeks. Figure~\ref{fig:interweek_correlations} presents the distribution of the coefficient over the commodities for lags from 1 to 10 weeks. It shows that there are substantial positive correlations between weeks for all commodities. Correlations are strong between successive weeks and are slightly weaker for longer lags. 

We now turn our attention to intercommodity correlations.
For each pair of commodities, we compute estimates of the Pearson correlation coefficient between weekly shipments over successive weeks. We present in Figure \ref{fig:intertype_commodities} the distribution of those correlations for lags from 1 to 10 weeks. There are 170 commodities, hence 28,730 intercommodity correlation coefficients to examine at each lag. Correlations vary across pairs of commodities. Some are large (positive or negative) for short and longer lags, while 50\% of the pairs have a weak correlation between -0.2 and 0.2. Outliers at each time lag represent 9\% of the significant coefficients. 

In summary, the data shows evidence of two types of correlation: strong positive interweek correlation between the weekly shipments over successive weeks, and strong intercommodity correlation between the weekly shipments of different commodities. Interweek correlation highlights the potential need for an autoregressive model while intercommodity correlation highlights a potential need for a model able to learn various dependence structures from the data. 
 
\subsubsection{Correlation Between Demand and Weather} \label{sec:corrWeather}

Weather can be an important aspect for rail freight transportation, especially in North America where railways extend over the subcontinent which is subject to major weather disruptions such as snowstorms. To assess the importance of weather on shipments, we use meteorological data and estimate the Pearson correlation coefficient between weekly shipments and several weather indicators. The data comes from National Oceanic and Atmospheric Administration (NOAA, https://www.ncdc.noaa.gov/) for terminals in the United States and from Statistics Canada (https://www.statcan.gc.ca) for terminals in Canada. More precisely, we use the average daily temperature and total daily snowfall in centimeters for the main 17 
terminals for the complete time range covered by the data. We compute the weekly temperature as the average temperature over the week and the accumulated snow (cm) as the sum of the daily values. 
For each terminal, at each week $t$, we sum the total departing and arriving demand over all commodities. 

\begin{figure}[htbp]
    \begin{subfigure}[b]{0.5\linewidth}
        \includegraphics[width=\linewidth]{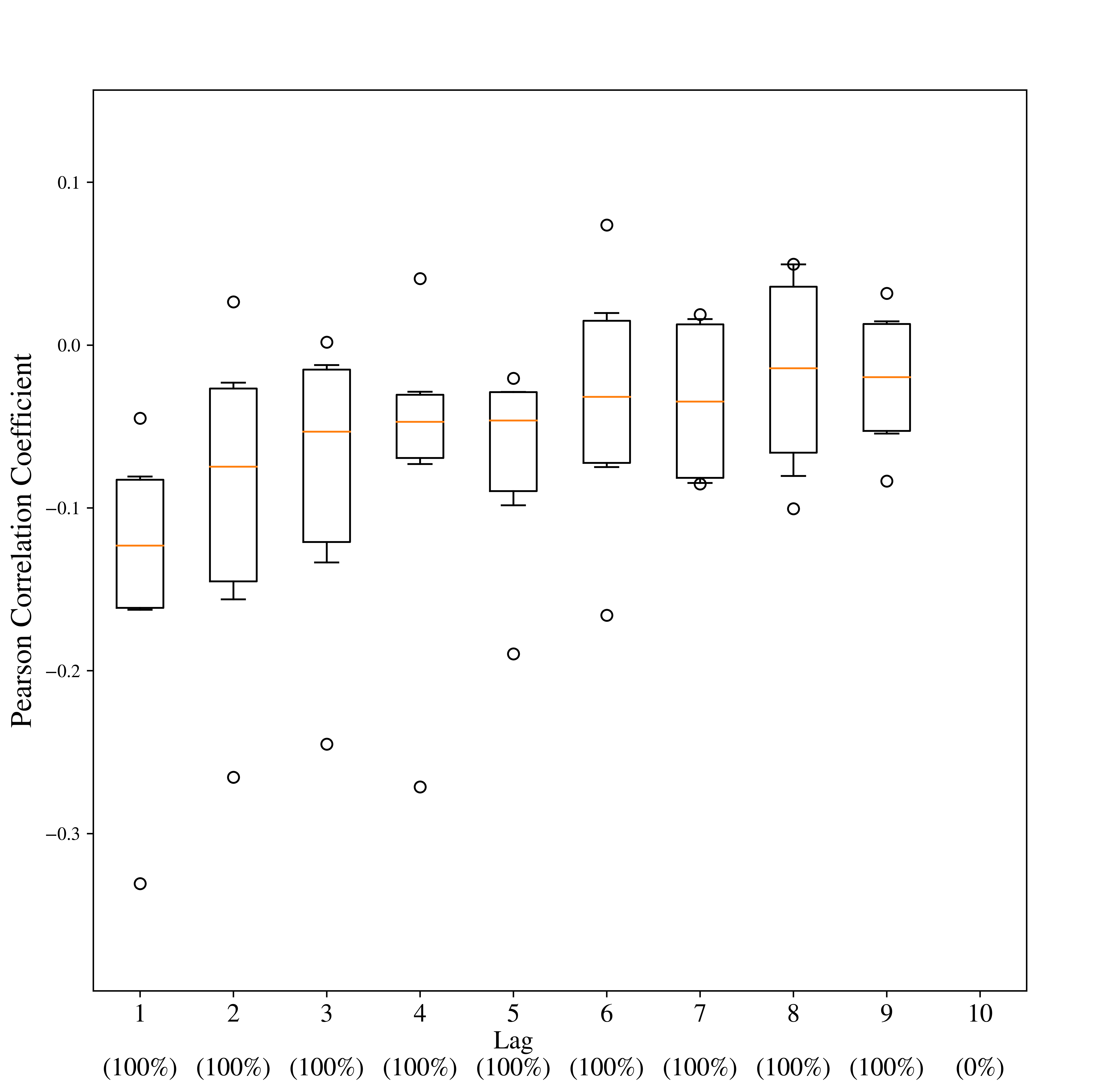}
        \caption{\tabular[t]{@{}l@{}} Departing Demand \endtabular}
        \label{fig:snow_departing}  
    \end{subfigure}
    \begin{subfigure}[b]{0.5\linewidth}
        \includegraphics[width=\linewidth]{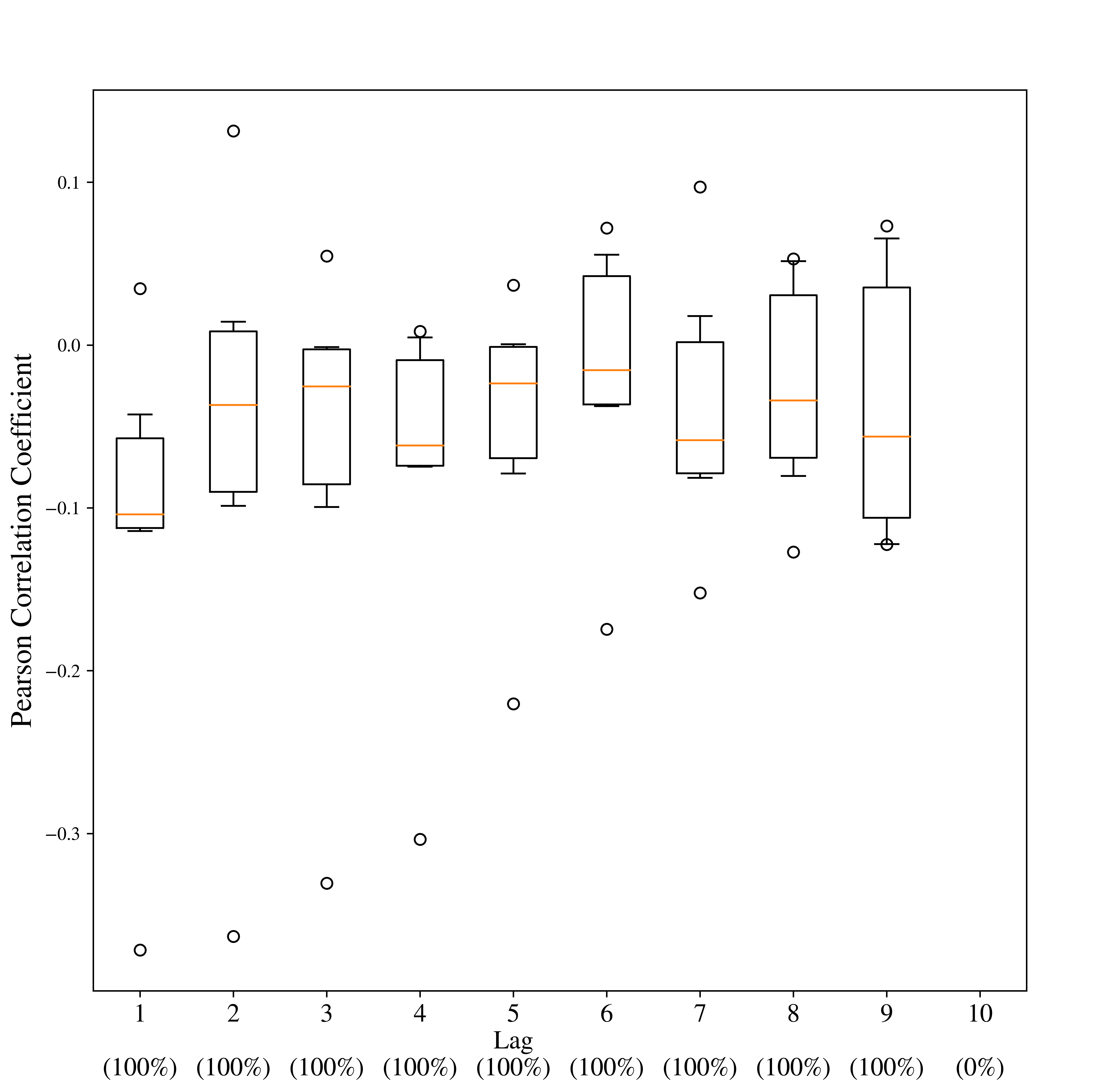}
        \caption{\tabular[t]{@{}l@{}} Arriving Demand \endtabular}
        \label{fig:snow_arriving}
    \end{subfigure}
\caption{Distribution of the Pearson correlation coefficient between accumulated snow over week $t$ and demand (arriving and departing) at week $t+\text{lag}$ over all the terminals. We consider only the coefficients which met the 95\% confidence level threshold and we indicate their proportion in parentheses. }
\label{fig:snow}
\end{figure}

Figure~\ref{fig:snow} shows the distribution over the terminals of the Pearson correlation coefficient between accumulated snow over week $t$ and departing and arriving demand at week $t+\text{lag}$. We note a negative correlation for both arriving and departing demand for successive weeks. It is weaker for longer lags.

\begin{figure}[htbp]
    \begin{subfigure}[b]{0.5\linewidth}
        \includegraphics[width=\linewidth]{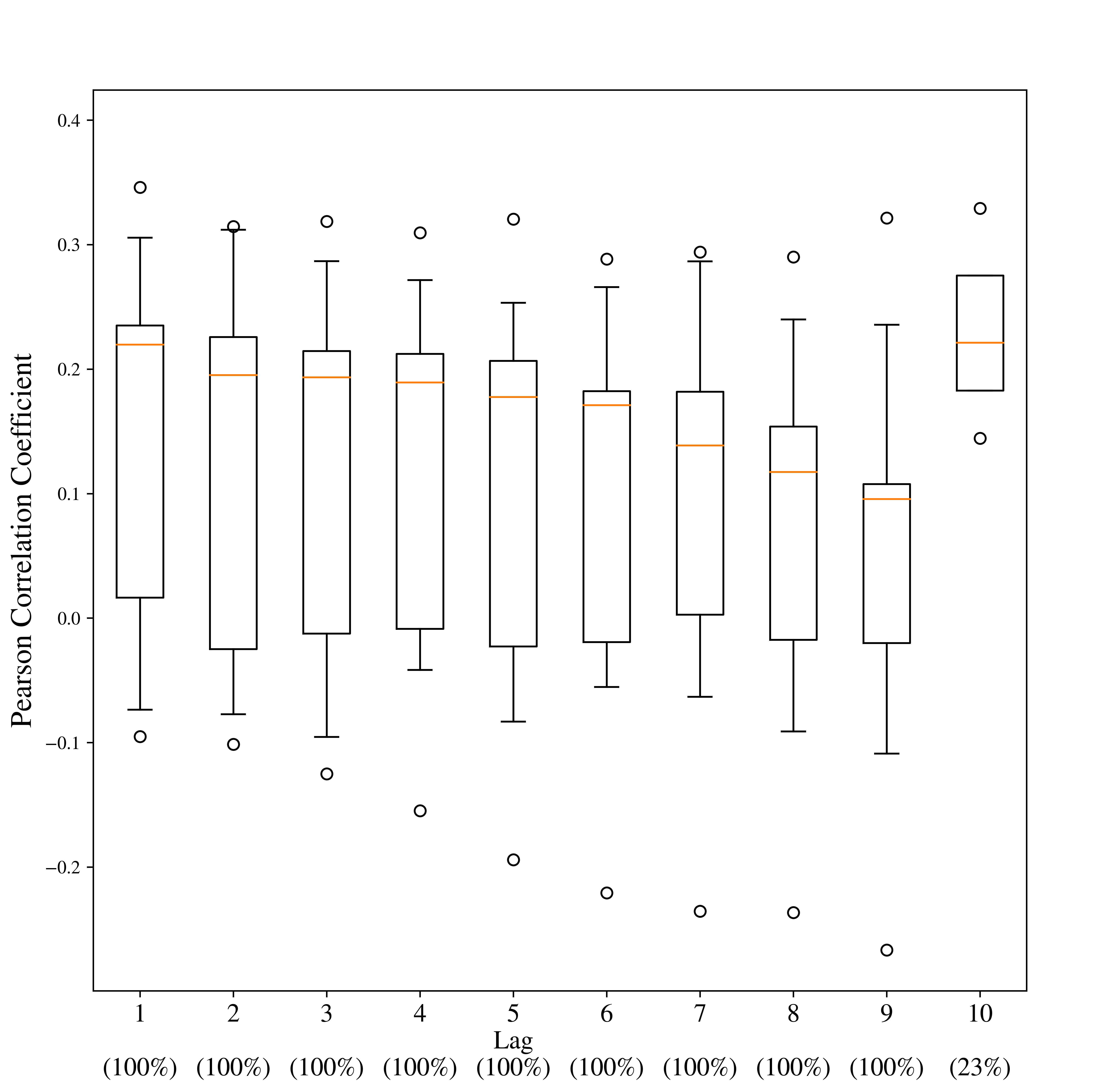}
        \caption{\tabular[t]{@{}l@{}} Departing Demand \endtabular}
        \label{fig:temperature_departing}  
    \end{subfigure}
    \begin{subfigure}[b]{0.5\linewidth}
        \includegraphics[width=\linewidth]{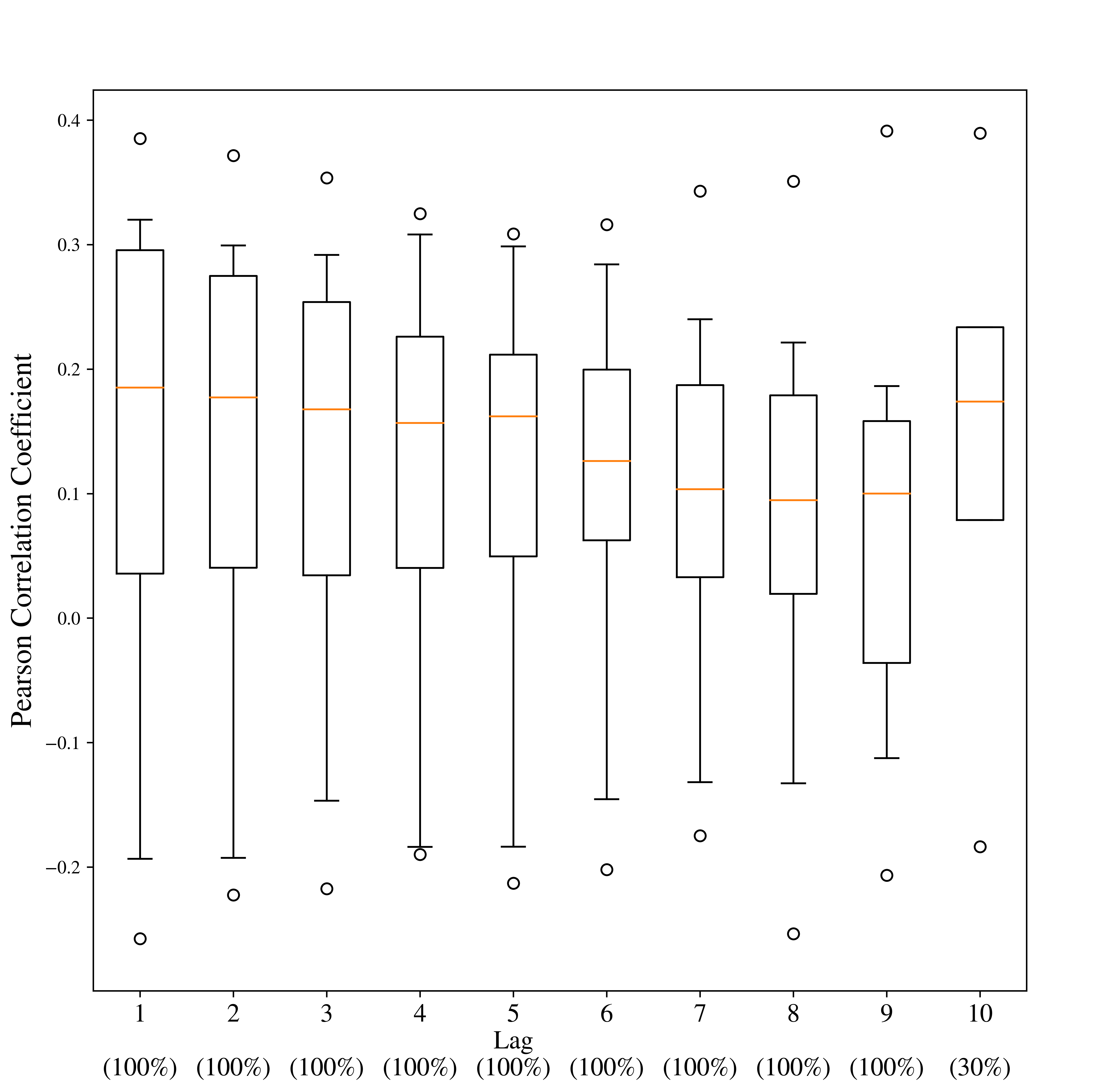}
        \caption{\tabular[t]{@{}l@{}} Arriving Demand \endtabular}
        \label{fig:temperature_arriving}
    \end{subfigure}
\caption{Distribution of the Pearson correlation coefficient between average temperature at week $t$ and demand (arriving and departing) at week $t+\text{lag}$ over all the terminals. We consider only the coefficients which met the 95\% confidence level threshold and we indicate their proportion in parentheses. }
\label{fig:temperature}
\end{figure}

We present in Figure~\ref{fig:temperature} the distribution over the terminals of the Pearson correlation coefficient between the average temperature and the demand arriving or departing over terminals. It shows an average positive correlation between demand and temperature which is weaker for longer lags. Some terminals have, however, negative correlations. This can be explained by a seasonality effect. On the one hand, summer and spring are busier periods for most ODs and temperatures are higher than the rest of the year. On the other hand, for import terminals, fall and the Chinese New Year are busier periods when temperatures are lower.

Correlation between demand and weather highlights the potential need for a forecasting model which can include weather features. In the following section, we present forecasting results for different forecasting models, ranging from very simple ones based on strong independence assumptions to neural networks that relax some of those assumptions.

\subsection{Time Series Forecasting Results}
\label{subsec:fcst_model_results}

\ef{Time series forecasts serve as input to the PDE problem. We therefore aim to assess the sensitivity with respect to this input. For this purpose we train models of different types: one statistical model (a basic autoregressive process, AR) and two machine learning models (a feed-forward neural network, FFNN, and a recurrent neural network, RNN). In the context of our application, there are advantages and drawbacks associated with each of the models. The purpose here is not to find the best possible forecasting model. Rather, we aim to assess the impact on the periodic demand estimates (hence the output of our methodology) of different types of models exhibiting different forecast error distributions. Based on the results reported in this section we retain two specific models that are used to solve the PDE problem (this is the focus of Section~\ref{subsec:bp_results}). Demand forecasting is known to be an inherently hard task with relatively large errors. In order to assess if this is also the case here, we include a naive baseline model (CONSTANT). If prediction errors are large for a naive model, and it is hard to improve upon this baseline, then this can be an indication that the prediction task is hard.}
%%We note that one has to be careful when comparing the different models as they do not all use the same features. 

\ef{In the following we provide some details on the data used for training and testing, and then we describe the models. Section~\ref{sec:forecastAccuracy} outlines our forecast accuracy measures and we report results in Section~\ref{sec:forecastResults}.}

We divide the dataset into a training, validation and test sets. Time series forecasting models require the last seen observed data before predicting the demand to come. Thus, we use the first \efsec{five} years of data for training (December 2013 - December 2018), the next \efsec{four} months of data for validation (January 2019 - April 2019) and the last \efsec{seven} months (May 2019 - November 2019) for testing. At each week $t_0$ in the dataset, we forecast demand for all commodities for $t = t_0+1, \dots ,t_0+T$. We consider $T=10$ weeks. To simplify the notation, we assume $t_0 = 0$ and refer to the estimates by $t = 1, \dots, T$. 

\paragraph{CONSTANT} This is the simplest possible model. It is based on the assumptions that commodities $k \in \mathcal{K}$ are independent and that the demand from observed week $t_0$ is the forecasts for the next $T$ weeks:
\begin{equation}
 \label{eq:naive}
    \hat{\mathbf{y}}_1 = \dots = \hat{\mathbf{y}}_T = \mathbf{y}_0.
\end{equation}

\paragraph{AR} For each commodity, we fit a \ef{basic} autoregressive $AR(p)$ process on the training data. This implies that the commodities are treated as independent. We use the estimated coefficients $\hat{\bm{\phi}} = (\hat{\phi}_1, \dots, \hat{\phi}_p)^T$ to compute the forecasts on the test set. The multistep forecasts for $t > 1$ are obtained using
\begin{equation}
    \hat{y}_{tk} = \hat{\phi}_{1k} \hat{y}_{t-1, k} + \dots + \hat{\phi}_{tk} y_{0k} + \dots + \hat{\phi}_{pk} y_{t-p, k}.
    \label{eq:multistep_ar}
\end{equation}

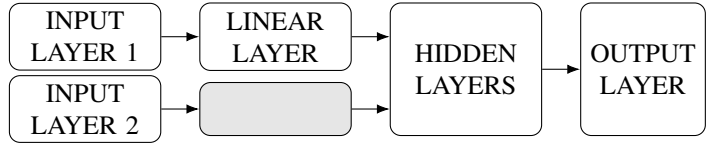
\begin{figure}[htbp]
\centering
\begin{tikzpicture}[ 
node distance = 0mm and 5mm,
    block/.style = {draw, rounded corners, minimum height=5em, text width=5em, align=center},
    block2/.style = {draw, rounded corners, minimum height=5em, text width=4em, align=center},
    halfblock/.style = {block, minimum height=2em},
    halfblock2/.style = {block, minimum height=2em, fill=black!10},
     line/.style = {draw, -Latex}
                       ]
% Place nodes
\node (n1) [block]  {HIDDEN LAYERS};
\node (n2) [block2, right=of n1] {OUTPUT \\ LAYER};  library
% input half boxes
\node (n1a) [halfblock, below left=of n1.north west]  {LINEAR LAYER};
\node (n1b) [halfblock2, above left=of n1.south west]  {};
% input layers
\node (n1a1) [halfblock, left=of n1a]  {INPUT LAYER 1};
\node (n1b1) [halfblock, left=of n1b]  {INPUT LAYER 2};
% arrows
\path [line]    (n1)  edge (n2)
                (n1a) edge (n1.west |- n1a)
                (n1b)  --  (n1.west |- n1b)
                (n1a1) edge (n1a)
                (n1b1) edge (n1b);
\end{tikzpicture}
\caption{Neural network architecture}
\label{fig:nn_draw}
\end{figure}

\paragraph{FFNN and RNN} We leverage the capacity of these models to forecast demand for multiple commodities simultaneously. As opposed to CONSTANT and AR, they relax the independence assumption on the commodities. We build one main neural network architecture described in Figure \ref{fig:nn_draw}. It is composed of 2 inputs layers, a stack of hidden dense or recurrent layers and one output layer. We consider several variants of this architecture. When the layers in the grey square are dense (or feed-forward), it forms the feed-forward architecture (FFNN). When the layers are LSTM, it forms the recurrent architecture (RNN). The dimension of the output layer is equal to the number of commodities that we predict simultaneously. 

At each period $t_0$, the output layer computes the forecast for the next time step $t_0+1$. Input Layer 1 is dedicated to external data such as weather features to model the correlation between demand and weather. Input Layer 2 is dedicated to the autoregressive modeling. It takes as input the demand of the previous weeks of all commodities, either observed or forecasted. When we generate the multistep demand forecasts, for Input Layer 2, we use $\hat{\mathbf{y}}_{t_0+1}$ to forecast demand at $t_0+2$, $\hat{\mathbf{y}}_{t_0+1}$ and $\hat{\mathbf{y}}_{t_0+2}$ for $t_0+3$, until $t_0+T$ with $\hat{\mathbf{y}}_{t_0+1}, \ldots, \hat{\mathbf{y}}_{t_0+T-1}$.

We evaluate several variants of the architecture to model the spatiotemporal correlations and the correlation with weather. They differ depending on the considered input data: lagged observed/forecasted demand and/or weather features. To train, validate and test the model, we use real observed weather data described in Section~\ref{subsec:data}. At prediction time, such information is not available and we would then rely on weather forecasts. These results are hence designed to assess the potential for weather-related features in an optimistic setting in regards to their accuracy. 

Forecasting 170 commodities simultaneously requires a rich and large dataset. Our dataset is fairly limited as it contains only 318 weekly demand data points for each commodity. To facilitate the forecasting task, we create a partition of $\mathcal{K}$ and train a neural network for each set of the partition. For this purpose, we split the set of commodities into 2 subsets: $\mathcal{K}_{53}$ which contains commodities of 53-feet container type and $\mathcal{K}_{40}$ which contains commodities of 40-feet container ($ |\mathcal{K}_{53}| = 58$ and $|\mathcal{K}_{40}| = 112$). 
Table~\ref{table:features_nn} summarizes the model variants.
Their names include a letter ``W'' if weather features are used to train the model, and ``SPLIT'' to indicate a partition of $\mathcal{K}$. 

\begin{table}[htbp]
\begin{tabular}{l|ccc}
 & Commodities & Weather features & Autoregressive Features \\ \hline
RNN           & $\mathcal{K}$   &  & \checkmark \\
FFNN           & $\mathcal{K}$   &  & \checkmark \\
RNN-W         & $\mathcal{K}$   & \checkmark & \checkmark \\
FFNN-W        & $\mathcal{K}$   & \checkmark & \checkmark \\
RNN-W-SPLIT1  & $\mathcal{K}_{53}$ & \checkmark & \checkmark \\
RNN-W-SPLIT2  & $\mathcal{K}_{40}$ & \checkmark & \checkmark \\
FFNN-W-SPLIT1 & $\mathcal{K}_{53}$ & \checkmark & \checkmark \\
FFNN-W-SPLIT2 & $\mathcal{K}_{40}$ & \checkmark & \checkmark
\end{tabular}
\caption{Features and set of commodities for each variant of the neural network architecture evaluated}
\label{table:features_nn}
\end{table}

The neural networks are trained with the backpropagation algorithm and the stochastic gradient descent using a Mean Squared Error (MSE) loss. For each model, we do a hyperparameter search with the Tree of Parzen Estimators implemented in the python library Hyperopt \citep{bergstra2013}. We select the set of hyperparameters which minimizes the MSE on the validation dataset. We report the detailed input features and the chosen set of hyperparameters for each trained model in the Appendix (Table~\ref{table:hyperparams}).

\subsubsection{Forecast Accuracy Measures} \label{sec:forecastAccuracy}
We measure the accuracy on the test set with two metrics: the Weighted Absolute Percentage Error (WAPE)
\begin{equation}
    \text{WAPE}_k = \frac{\sum_{t_0 \in \mathcal{D}_{\text{test}}} \sum_{t = 1}^T |y_{t_0+t,k} - \hat{y}_{t_0+t,k}|}{\sum_{t_0 \in \mathcal{D}_{\text{test}}} \sum_{t = 1}^T y_{t_0+t,k}} \times 100, ~k=1,\ldots,K,
    \label{eq:wape}
\end{equation}
where $\mathcal{D}_{\text{test}}$ is the test set of size $N$, and the Root Mean Squared Error (RMSE)
\begin{equation}
    \text{RMSE}_k = \sqrt{ \frac{1}{N \times T} \sum_{t_0 \in \mathcal{D}_{\text{test}}} \sum_{t=1}^T \left( y_{t_0+t,k} - \hat{y}_{t_0+t,k} \right)^2 }, ~k=1,\ldots,K.
    \label{eq:rmse}
\end{equation}

The WAPE is a weighted version of the Mean Absolute Percentage Error (MAPE) which handles small or zero demand values. Low demands are frequent in our data for some commodities with sparse demand over the year. Hence the importance of having a metric independent to the scale of the time series such as WAPE. The RMSE \greta{is also an important metric to consider, as it} puts a high weight on large errors.

\subsubsection{Results} \label{sec:forecastResults}

Table~\ref{table:avg_results_fcst} reports the performance metrics averaged over all commodities. We note that the metrics for models based on a partition of the commodities (SPLIT) are averaged over the partitions. 
The results show that the AR \greta{model} has the best performance and it is considerably better than the NN models. The CONSTANT baseline has a WAPE close to AR but a considerably worse RMSE. \ef{The fact that the errors are overall large and that the performance of CONSTANT is close to the other models in terms of WAPE, is an indication that the prediction task is hard.} 

The descriptive statistics in Section~\ref{sec:correlationComm} show strong intercommodity correlations. Nevertheless, the AR model, based on the assumption that demands for commodities are independent, performs better than the NN models where this assumption is relaxed. Neural networks have more parameters to fit on the same limited data. While they have the capacity to model non-linear relationships, they also require more data to be trained. We believe that our data containing only 318 observations for each commodity is too limited for training the NN models. We note that these findings are consistent with other studies \citep[e.g.,][]{makridakis2018statistical}. That is, basic time series models outperform deep learning models on difficult time series forecasting tasks, such as this one.  \ef{Recall that we aggregate the daily data into weekly observations to circumvent issues related to demand being constrained by the supply (data truncation and censoring). Without the aggregation we would have had seven times more observations. However, even 2,226 observations is considered a very small dataset for training neural networks and daily observations have higher variance than weekly ones.}

\begin{table}[htbp]
\centering
\begin{tabular}{l|rr}
{Model} & {RMSE} & {WAPE} \\ \hline
CONSTANT & 86.0 & \textbf{34.7\%} \\
AR & \textbf{78.0} & \textbf{34.0\%} \\
FFNN & 105.2 & 38.7\% \\
FFNN-W & \textbf{84.8} & 37.1\% \\
FFNN-W-SPLIT & 90.4 & 37.2\% \\
RNN & 105.1 & 37.8\% \\
RNN-W & 86.3 & 37.8\% \\
RNN-W-SPLIT & 85.2 & 38.6\%
\end{tabular}
\caption{Performance metrics of the forecasting models averaged over all commodities. The best and second-best metric values are highlighted in bold.}
\label{table:avg_results_fcst}
\end{table}

The results for the deep learning models confirm that adding weather features and considering all commodities simultaneously help to improve the performance. This is consistent with the descriptive statistics reported in Section~\ref{sec:corrWeather}. Previous demands for all commodities constitute relevant information to consider. 

Henceforth, we keep two forecasting models when analyzing periodic demand results: the overall best performing model (AR) as well as the best deep learning model (FFNN-W).

\subsection{Periodic Demand Estimation}
\label{subsec:bp_results}

\greta{Before diving into the results for the periodic demand estimation problem, we recall some important details. In the \textbf{PDE} program \eqref{eq:objP1}-\eqref{eq:last_P}, the decision variables relate to the periodic demand estimate $\hat{\mathbf{y}}^{\text{p}}$, defined as a mapping from the demand forecasts. It serves as a parameter of the SND problem at the second level, \textbf{BP}, defining the services offered by the carrier at each period of the tactical horizon.
The third level \textbf{wBP} corresponds to the flow problem with fixed services but where the demand (and hence flow variables) differs from the periodic demand estimate.
The matrix of demand forecasts, $\hat{\mathbf{Y}}$ is an input to \textbf{wBP}. 
Our objective is to analyze the impact of the periodic demand estimate. That is, the costs resulting from selecting a specific $\hat{\mathbf{y}}^{\text{p}}$ to solve \textbf{BP} and \textbf{wBP}. We therefore} divide the results into two parts. The purpose is to disentangle the errors associated with the periodic demand estimation from those associated with the demand forecasts. The two parts are briefly described in the following:
\begin{itemize}
    \item \textbf{Analysis~1: \ef{T}he impact of periodic demand estimation}. We assume we have no forecast errors, i.e., the carrier knows perfectly the demand to come for the planning horizon. The periodic demand is estimated with the mappings~\eqref{eq:max_periodic}-\eqref{eq:q3_periodic} from historical data (ground truth values) and we compute the tactical costs generated by those periodic demands.
    \item \textbf{Analysis~2: \ef{T}he impact of imperfect demand forecasts}. We estimate the periodic demands from forecasts obtained with the AR and FFNN-W models. We compute the associated tactical costs and compare them to Analysis~1.
\end{itemize}

\begin{figure}
\centering
\includegraphics[width=\linewidth]{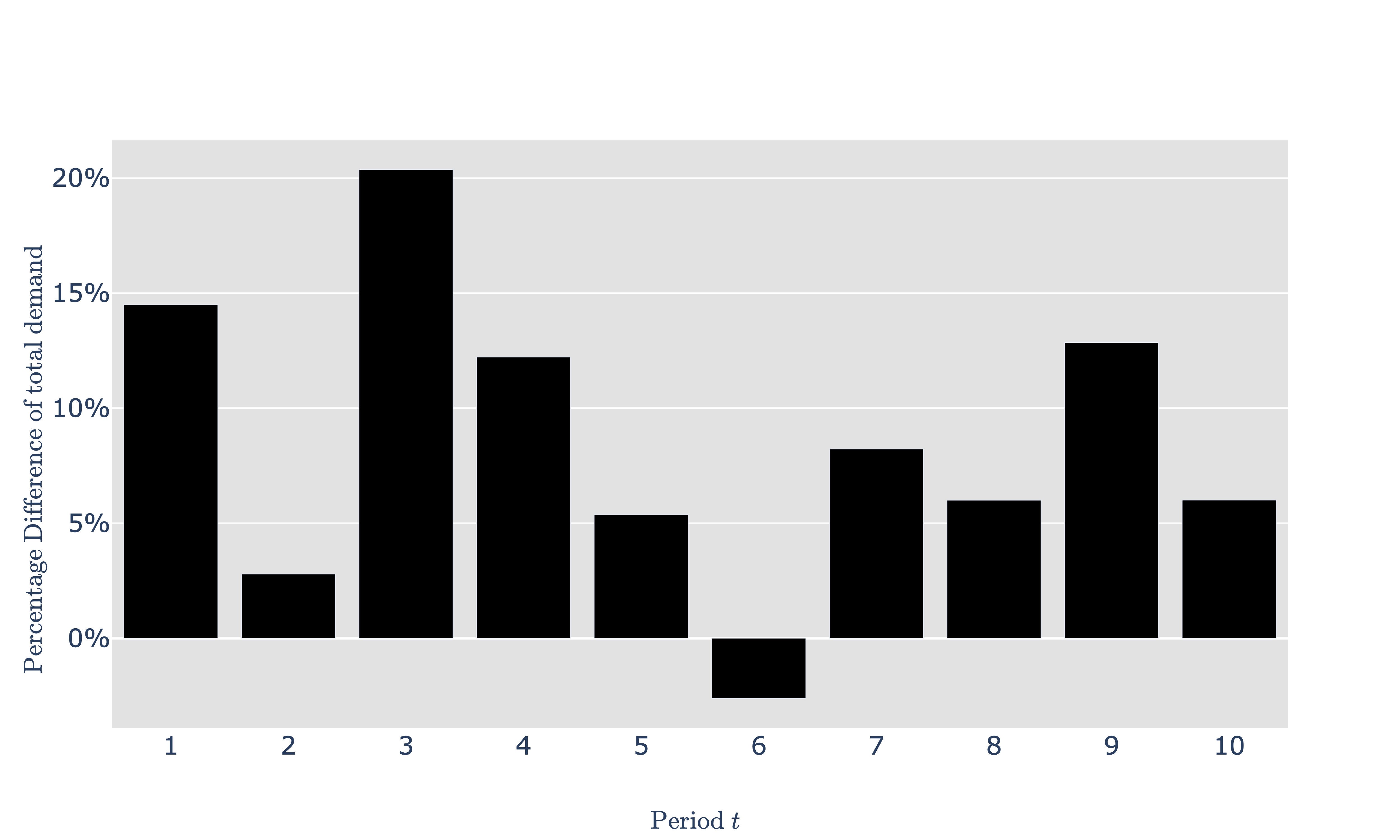}
\caption{Percentage difference of the total demand at each week of $I2$ relative to  $I1$}
\label{fig:I2_vs_I1}
\end{figure}

We consider two demand instances -- $I1$ and $I2$ -- from the test set of the forecasting models. They are from two distinct periods: $I1$ \ef{covers a spring/summer period}
\ef{(}May 6th to July 14th, 2019\ef{) while $I2$} 
\ef{covers a summer/fall period}
\ef{(}July 29th to October 6th, 2019\ef{)}. Both instances have $K=170$ commodities and a tactical planning horizon of $T=10$ weeks. Instance $I2$ corresponds to a busier period for the carrier: Figure~\ref{fig:I2_vs_I1} shows the difference of the total demand summed over all commodities of $I2$ at each week relative to the total demand of $I1$. We note that the total demand for $I2$ can be up to 20\% higher than its $I1$ counterpart. 

All the results are generated by fixing
the variable $\hat{\mathbf{y}}^{\text{p}}$ in \textbf{PDE} and sequentially solving \textbf{BP-wBP}. We recall that the periodic demand $\mathbf{y}^{\text{p}}$ is a mapping from \ef{actual} demand values \ef{while}  $\hat{\mathbf{y}}^{\text{p}}$ is a mapping from demand forecasts\ef{. Similarly,} $\mathbf{Y}$ is the matrix of \ef{actual} demand values \ef{while} $\hat{\mathbf{Y}}$ is the matrix of demand forecasts. We denote by  $\textbf{BP}(\mathbf{y}^\text{p})$ and $\textbf{BP}(\hat{\mathbf{y}}^\text{p})$ when \textbf{BP} is solved with $\mathbf{y}^\text{p}$ and $\hat{\mathbf{y}}^\text{p}$, respectively, in constraints~\eqref{eq:periodic_bp}. Furthermore, we denote by $\textbf{wBP}(\mathbf{Y})$ and $\textbf{wBP}(\hat{\mathbf{Y}})$ when \textbf{wBP} is solved with demand values $\mathbf{Y}$ and $\hat{\mathbf{Y}}$, respectively, in constraints~\eqref{eq:pde_wbp_2}.
\gl{Finally, due to the large number of variables (several million) of \eqref{eq:objP1}-\eqref{eq:last_P} for both instances $I1$ and $I2$, we focus on a comprehensive analysis of aggregate statistics of solution quality, rather than values of decision variables.} %Indeed, looking only at decision variables, there are 170 variables $y^{\text{p}}$, 2,208 variables $z$ and 4,128,960 variables $x$}. 

\subsubsection{Analysis 1: The Impact of Periodic Demand Estimation}
The analysis in this section is based on two sets of results for demand instances $I1$ and $I2$:

\begin{itemize}
    \item \textbf{Reference}: We solve $\textbf{BP}(\mathbf{y}_t)$ for $t=1,\ldots, T$. In other words, we do not restrict the demand to be periodic and we obtain a tactical cost $C_{\text{ref}}^{\text{PDE}}$~\eqref{eq:objP1}.  We note that this constitutes a lower bound on $C^{\text{PDE}}$: At each week, the blocks built are specific for the demand to best exploit the network capacity.
    \item \textbf{Periodic}: We solve  $\textbf{BP}(\mathbf{y}_{.}^{\text{p}})-\textbf{wBP}(\mathbf{Y})$ with the four periodic demand vectors $\mathbf{y}^{\text{p}}_{\text{mean}}$, $\mathbf{y}^{\text{p}}_{\text{max}}$, $\mathbf{y}^{\text{p}}_{\text{q2}}$ and $\mathbf{y}^{\text{p}}_{\text{q3}}$. For each solution, we compute the percentage gap to the reference cost $C_{\text{ref}}^{\text{PDE}}$.
\end{itemize}

Table~\ref{table:comparison_actuals_ref} reports the results where each number is the percentage gap to the reference cost $C_{\text{ref}}^{\text{PDE}}$. \efsec{Since the latter is a lower bound, column $C^{\text{PDE}}$ contains only positive values.} 
\efsec{Recall that costs at an aggregate level -- $C^{\text{design}}$, $C^{\text{flow}}$ and $C^{\text{out}}$ -- result from non-trivial trade-offs between costs associated with design and flow decisions at the path level that, in turn, are all linked through the demand.} Hence, we cannot analyze each component independently of the others. \efsec{We can, however, note a trend due to the nature of periodic demand. Namely, $C^{\text{design}}$ and $C^{\text{flow}}$ decrease while $C^{\text{out}}$ increases when periodic demand is used since the plan then needs to be adapted to the weekly demand. If actual weekly demand cannot be satisfied by the periodic plan, then it will be outsourced at a per unit flow cost $C^{\text{out}}_{tbk}$ that is higher than that of the periodic plan $C_{tbk}^{\text{flow}}$.}
%
%\ef{Recall that} the cost $C^{\text{PDE}}$ \ef{constitutes} a non-trivial trade-off between \ef{its} three \ef{terms} $C^{\text{design}}$, $C^{\text{flow}}$ and $C^{\text{out}}$, that, \ef{in turn}, are \ef{all} linked \ef{to the} demand. Hence, we cannot analyze each component independently of the others. 
The parameter values in the objective function~\eqref{eq:objP1} have been \ef{fixed through an extensive validation process} with CN \ef{as part of the work reported in \cite{Morganti2019}. The values are such that they represent actual costs whenever possible and, in particular, that the solutions reflect CN's operations and are implementable in practice. Therefore, we treat the parameter values as fixed and given in this study.} 

\begin{table}[htbp]
\centering
\begin{tabular}{ccc:ccc}
 & \multicolumn{1}{c}{\multirow{2}{*}{\begin{tabular}[c]{@{}c@{}}Periodic \\ Definition $\mathbf{y}^{\text{p}}$ \\ in $\textbf{BP}$ \end{tabular}}} & \multicolumn{4}{c}{Percentage \gl{gap between} costs} \\
 & \multicolumn{1}{c}{} & \makecell{$C^{\text{PDE}}$\\} & \makecell{$C^{\text{design}}$ \\} & \makecell{$C^{\text{flow}}$ \\ } & \makecell{$C^{\text{out}}$ \\ } \\ 
%  &  &  &  &  & \\ 
 &  &  &  &  &  \\ \hline
 &  &  &  &  & \\ 
     & $\mathbf{y}^{\text{p}}_{\text{max}}$  & 77.3\% & -2.8\%  &  -3.4\% & 84.0\% \\ 
$I1$ & $\mathbf{y}^{\text{p}}_{\text{mean}}$ & 93.3\% & -16.9\% &  -4.4\% & 101.3\% \\
     & $\mathbf{y}^{\text{p}}_{\text{q2}}$   & 93.8\% & -16.8\% &  -3.3\% & 101.8\% \\
     & $\mathbf{y}^{\text{p}}_{\text{q3}}$   & \textbf{28.4\%} & -7.5\%  &  -1.2\% & 30.8\%  \\
 &  &  &  &  &  \\ \hline
 &  &  &  &  &  \\
     & $\mathbf{y}^{\text{p}}_{\text{max}}$  & 73.0\% & -14.1\% & -5.7\% & 76.6\% \\
$I2$ & $\mathbf{y}^{\text{p}}_{\text{mean}}$ & 40.6\% & -20.0\% & -2.8\% & 42.6\% \\
     & $\mathbf{y}^{\text{p}}_{\text{q2}}$   & 41.6\% & -23.2\% & -2.8\% & 43.6\% \\
     & $\mathbf{y}^{\text{p}}_{\text{q3}}$   & \textbf{23.1\%} & -14.8\% & -2.6\% & 24.3\% \\
\end{tabular}
\caption{\efsec{Percentage gap between tactical cost $C^{\text{PDE}}$} and reference cost $C_{\text{ref}}^{\text{PDE}}$. We use historical data for the demand at each week, that is $\textbf{wBP}(\mathbf{Y})$. The minimum value for $C^{\text{PDE}}$ indicates the optimal periodic demand estimate in the subset $\mathcal{Y}$ we consider.}
\label{table:comparison_actuals_ref}
\end{table}

Two important findings emerge \ef{from the results}. First, the tactical cost has an important variation over the different periodic demand estimates. This underlines the importance of the periodic demand estimation problem. Second, using another estimate than the commonly used mean periodic demand $\mathbf{y}^{\text{p}}_{\text{mean}}$ can lead to an important cost reduction. In our case using the third quartile $\mathbf{y}^{\text{p}}_{\text{q3}}$ reduces the total costs by 33.6\% in $I1$, and by 12.4\% in $I2$ with respect to $\mathbf{y}^{\text{p}}_{\text{mean}}$. When a smaller periodic demand estimate is used, $\mathbf{y}^{\text{p}}_{\text{mean}}$ and $\mathbf{y}^{\text{p}}_{\text{q2}}$ for instance, less blocks are built \gl{(the numbers of variables where $z_b = 1$ decreases)}\efsec{. This leads to increased outsourced demand when actual demand is higher than planned as there are not enough blocks for all commodities.} %at the expense of outsourced demand. 

We analyze total periodic demand \efsec{($T \sum_{k \in \mathcal{K}} y_k^{\text{p}}$)} relative to total actual demand \efsec{($\sum_{t=1}^T\sum_{k \in \mathcal{K}} y_{tk}$)} to explain the results. We report the values \efsec{for each mapping} in  Table~\ref{table:relative_demands_reference}. The total periodic demand of $\mathbf{y}^\text{p}_{\text{max}}$ is almost three times larger than the one of $\mathbf{y}^{\text{p}}_{\text{q3}}$ in both instances which explains the large gap in their costs. The design made from solving $\textbf{BP}(\textbf{y}_{\text{q3}}^{\text{p}})$ is based on a lower estimation of the periodic demand than $\textbf{BP}(\textbf{y}_{\text{max}}^{\text{p}})$ which generates fewer blocks in the network. The advantage
is that they \efsec{can be} better used\efsec{. T}he design from solving $\textbf{BP}(\textbf{y}_{\text{max}}^{\text{p}})$ is made of a large number of small blocks \gl{(the number of variables where $z_b = 1$ is large)} % but the number of variables where $x_{tbk} > 0$ decreases)}  
which cannot accommodate all actual demands. \efsec{In this case, outsourced demand increases for certain commodities because the shared train capacities are divided between a large number of blocks (some actual demands $x_{tbk}$ need to be fixed to zero to satisfy capacity constraints).}
%Hence the increase in outsourced demand. 
We highlight that while the increase in outsourced demand seems large, it represents only few percent of the total demand.

\begin{table}[htbp]
\centering
\begin{tabular}{lc|c}
 & $I1$ & $I2$ \\ \hline
$\sum_{t=1}^T\sum_{k \in \mathcal{K}} y_{tk}$ & - & - \\
$\mathbf{y}^\text{p}_{\text{max}}$ & 34.8\% & 31.8\% \\
$\mathbf{y}^{\text{p}}_{\text{mean}}$ & 0.0\% & 0.0\% \\
$\mathbf{y}^\text{p}_{\text{q2}}$ & -1.0\% & -0.5\% \\
$\mathbf{y}^{\text{p}}_{\text{q3}}$ & 11.8\% & 11.7\%
\end{tabular}
\caption{\efsec{Total periodic demand relative to total actual demand.} The point of reference is indicated by a dash. }
\label{table:relative_demands_reference}
\end{table}

% ************ IMPORTANT ************
% Instance 1 = H2
% Instance 2 = H14
% Instance 3 = H20
% Instance 4 = H11
% Instance 5 = H17
% ******************************

\subsubsection{Analysis 2: The Impact of Imperfect Demand Forecasts}

In practice, carriers rely on demand forecasts to estimate a periodic demand and build the design for the tactical planning horizon. Then, at each week of the horizon, they adapt operationally the tactical plan based on observed demand. In this section, we follow our methodology and analyze the quality of the solution a posteriori. 

We proceed in two steps: First, we estimate the periodic demand by solving $\textbf{BP}(\hat{\mathbf{y}}^{\text{p}})-\textbf{wBP}(\hat{\mathbf{Y}})$ \ef{(i.e., using the forecasts)} for each mapping $h$ and select the one minimizing $C^{\text{PDE}}$~\eqref{eq:objP1}. Second, we assess the tactical cost associated with the periodic demand estimate. For this purpose we solve $\textbf{BP}(\hat{\mathbf{y}}^{\text{p}})-\textbf{wBP}(\mathbf{Y})$. Note that we use \ef{actual} demand values $\mathbf{Y}$ to accurately assess this cost. Hence, it is affected by two combined sources of error: the one of the periodic demand estimation and the forecast error, both discussed separately in previous sections.

\paragraph{Step 1: Periodic demand estimation.} We report results in Table~\ref{table:solving_PDE_forecasts} and indicate by a dash the point of reference ($\hat{\mathbf{y}}^{\text{p}}_{\text{mean}}$). We note that \ef{the forecasts} $\hat{\mathbf{Y}}$ depend on both the forecasting model and the instance. In other words, we can only compare results from the same forecasting model and the same instance. The results show that, \ef{in all cases}, the tactical costs are minimized with $\hat{\mathbf{y}}_{\text{max}}^{\text{p}}$. \ef{In contrast,} Analysis~1 \ef{based on} historical data \ef{shows that} tactical costs \ef{are} minimized with ${\mathbf{y}}_{\text{q3}}^{\text{p}}$. \ef{This can be explained by} the fact that forecasting models smooth demand and struggle to accurately predict  peaks. \ef{With the} historical data we have access to \greta{the} maximum demand which can be an outlier, thus expensive. 
Following the methodology, we choose the periodic demand \ef{estimate} having the lowest cost for Step 2\ef{. It is $\hat{\mathbf{y}}_{\text{max}}^{\text{p}}$} for both forecasting models.

\begin{table}[htbp]
\centering
\begin{tabular}{ccccc}
 &  &  & \multicolumn{2}{c}{Forecasting Model} \\ %\cline{4-5} 
 %&  &  &  &  \\
 & \makecell{Periodic \\ Demand $\mathbf{y}^{\text{p}}$ \\in \textbf{BP}} &   & \multicolumn{1}{c|}{AR} & FFNN-W \\ \Xhline{3\arrayrulewidth}
 &  &  & \multicolumn{1}{c|}{} &  \\
 & $\hat{\mathbf{y}}^{\text{p}}_{\text{max}}$ & & \multicolumn{1}{c|}{\textbf{-38.4\%}} & \textbf{-32.2\%} \\
$I1$ & $\hat{\mathbf{y}}^{\text{p}}_{\text{mean}}$ &  & \multicolumn{1}{c|}{-} &  -\\
 & $\hat{\mathbf{y}}^{\text{p}}_{\text{q2}}$ &  & \multicolumn{1}{c|}{17.5\%} & 28.0\% \\
 & $\hat{\mathbf{y}}^{\text{p}}_{\text{q3}}$ &  & \multicolumn{1}{c|}{-30.8\%} &  -18.8\%\\
 &  &  & \multicolumn{1}{c|}{} &  \\ \Xhline{3\arrayrulewidth}
 &  &  & \multicolumn{1}{c|}{} &  \\
& $\hat{\mathbf{y}}^{\text{p}}_{\text{max}}$ &  & \multicolumn{1}{c|}{\textbf{-7.4\%}} & \textbf{-28.4\%} \\
$I2$ & $\hat{\mathbf{y}}^{\text{p}}_{\text{mean}}$ &   & \multicolumn{1}{c|}{-} & - \\
& $\hat{\mathbf{y}}^{\text{p}}_{\text{q2}}$ &   & \multicolumn{1}{c|}{12.7\%} & 29.2\% \\
& $\hat{\mathbf{y}}^{\text{p}}_{\text{q3}}$ &   & \multicolumn{1}{c|}{8.8\%} & -11.7\%
\end{tabular}
\caption{Percentage \gl{gap between} tactical costs $C^{\text{PDE}}$ resulting from solving $\textbf{BP}(\hat{\mathbf{y}}^{\text{p}})-\textbf{wBP}(\hat{\mathbf{Y}})$ with forecasts of demand from two models: AR and FFNN-W. For each model and each instance, we compare the value with the one from $\hat{\mathbf{y}}^{\text{p}}_{\text{mean}}$.}
\label{table:solving_PDE_forecasts}
\end{table}

\paragraph{Step 2: Assessment of the tactical costs\efsec{.}} Table~\ref{table:total_costs_relative_v2} reports the relative costs resulting from solving $\textbf{BP}(\hat{\mathbf{y}}^{\text{p}})-\textbf{wBP}(\mathbf{Y})$. We use the same reference as in Analysis~1, that is $C_{\text{ref}}^{\text{PDE}}$, the lower bound on $C^{\text{PDE}}$. In addition to the best periodic demand identified by our methodology, we report the result for $\hat{\mathbf{y}}^{\text{p}}_{\text{mean}}$. Consistent with the findings in Analysis~1, we note that using the latter leads to a large increase in costs. Despite the relatively large forecast errors reported in Table~\ref{table:avg_results_fcst}, the results show that estimating periodic demand using our methodology can even reduce the costs compared to $\mathbf{y}^{\text{p}}_{\text{mean}}$ computed on \emph{perfect information} (historical data).

\begin{table}[htbp]
\centering
\begin{tabular}{cccc:ccc}
&  & \multicolumn{1}{c}{\multirow{2}{*}{\begin{tabular}[c]{@{}c@{}}Periodic \\ Demand $\mathbf{y}^{\text{p}}$ \\ in $\textbf{BP}$ \end{tabular}}} & \multicolumn{4}{c}{Percentage \gl{gap between} costs} \\
&  & \multicolumn{1}{c}{} & \makecell{$C^{\text{PDE}}$\\} & \makecell{$C^{\text{design}}$ \\} & \makecell{$C^{\text{flow}}$ \\ } & \makecell{$C^{\text{out}}$ \\ } \\ 
%  &  &  &  &  & \\ 
&  &  &  &  &  &  \\ \hline
&  &  &  &  &  & \\ 
     & Historical data  &   $\mathbf{y}^{\text{p}}_{\text{mean}}$  & 93.3\%  & -16.9\% & -4.4\% & 101.3\% \\
     & Historical data  &	$\mathbf{y}^{\text{p}}_{\text{q3}}$	  & 28.4\%  & -7.5 \% & -1.2\% & 30.8\% \\
$I1$ & AR               & 	$\hat{\mathbf{y}}^{\text{p}}_{\text{max}}$   & 55.9\%  & -17.8\% & -1.9\% & 60.6\% \\
     & AR               & 	$\hat{\mathbf{y}}^{\text{p}}_{\text{mean}}$  & 125.1\% & -14.1\% & -3.9\% & 135.8\% \\
     & FFNN-W           & 	$\hat{\mathbf{y}}^{\text{p}}_{\text{max}}$   & 48.2\%  & -12.3\% & -2.5\% & 52.4\% \\
     & FFNN-W           & 	$\hat{\mathbf{y}}^{\text{p}}_{\text{mean}}$  & 117.4\% & -16.2\% & -6.0\% & 127.5\% \\
&  &  &  &  &  &  \\ \hline
&  &  &  &  &  &  \\
     & Historical data  &	$\mathbf{y}^{\text{p}}_{\text{mean}}$	     & 40.6\%  & -20.0\% & -2.8 \% & 42.6\%\\
     & Historical data  &	$\mathbf{y}^{\text{p}}_{\text{q3}}$	         & 23.1\%  & -14.8\% & -2.6 \% & 24.3\%\\
$I2$ & AR               & 	$\hat{\mathbf{y}}^{\text{p}}_{\text{max}}$   & 50.1\%  & -23.6\% & -10.0\% & 52.8\%\\
     & AR               & 	$\hat{\mathbf{y}}^{\text{p}}_{\text{mean}}$  & 137.6\% & -31.8\% & -12.1\% & 144.4\%\\
     & FFNN-W           & 	$\hat{\mathbf{y}}^{\text{p}}_{\text{max}}$   & 80.0\%  & -25.3\% & -7.1 \% & 83.9\%\\
     & FFNN-W           & 	$\hat{\mathbf{y}}^{\text{p}}_{\text{mean}}$  & 164.4\% & -31.1\% & -17.8\% & 172.6\% \\
\end{tabular}
\caption{Percentage \gl{gap between} tactical costs resulting from solving $\textbf{BP}(\hat{\mathbf{y}}^{\text{p}})-\textbf{wBP}(\mathbf{Y})$ \gl{and reference cost $C_{\text{ref}}^{\text{PDE}}$ }%. The point of reference is $C_{\text{ref}}^{\text{PDE}}$, the reference 
used in Analysis~1 ($\textbf{BP}(\mathbf{y}_t)$ for $t=1,\ldots, T$). }
\label{table:total_costs_relative_v2}
\end{table}

%Percentage gap between tactical cost $C^{\text{PDE}}$} and reference cost $C_{\text{ref}}^{\text{PDE}}$.

We report the \gl{relative} total demand values \efsec{($T \sum_{k \in \mathcal{K}} y_k^{\text{p}}$)} in Table~\ref{table:relative_demands}. For  instance $I1$, both periodic demand \ef{estimates} $\hat{\mathbf{y}}^{\text{p}}_{\text{max}}$ overestimate the reference demand. The design built is capable of handling more demand, which results in less outsourced demand. \greta{However, they underestimate the periodic demand $\mathbf{y}^{\text{p}}_{\text{q3}}$ (based on actual demand) hence the plan is built on a reduced demand, which in turn lead to more outsourced demand compared to $\mathbf{y}^{\text{p}}_{\text{q3}}$.}
\ef{For} instance $I2$, the periodic \ef{demand estimate} $\hat{\mathbf{y}}^{\text{p,AR}}_{\text{max}}$ overestimates the total demand yet it leads to an increase in outsourced demand. This is because \ef{the estimate} $\hat{\mathbf{y}}^{\text{p,AR}}_{\text{max}}$ either overestimates demand for commodities that already lack capacity with \ef{actual average demand} $\mathbf{y}^{\text{p}}_{\text{mean}}$, or it underestimates demand for some  commodities for which blocks are then not built in the design and are consequently outsourced at each week.

\begin{table}[htbp]
\centering
\begin{tabular}{llc|c}
&  & $I1$ & $I2$ \\ \hline
Historical data & $\mathbf{y}^{\text{p}}_{\text{mean}}$ & - & - \\
Historical data & $\mathbf{y}^{\text{p}}_{\text{q3}}$ & 11.9\% & 11.7\% \\
AR & $\hat{\mathbf{y}}^{\text{p}}_{\text{max}}$ & 10.2\% & 0.7\% \\
FFNN-W & $\hat{\mathbf{y}}^{\text{p}}_{\text{max}}$ & 10.0\% & -1.4\%
\end{tabular}
\caption{Total periodic demand summed over commodities. The point of reference is indicated by a dash. }
\label{table:relative_demands}
\end{table}

\ef{In summary, the results show that the PDE problem can be of high value. Based on ground truth data (no forecast error) we obtain an important cost reduction using the third quartile as opposed to the mean over the tactical planning horizon. Even though our time series forecasts have relatively high errors, the periodic demand estimate resulting from our methodology leads to a cost reduction compared to the commonly used average value computed based on \emph{ground truth values}. This further highlights the importance of the problem introduced in this paper. }

\section{Conclusion and Future Research}
\label{section:ccl}

Tactical planning is essential to freight carriers. \ef{For example, it aims at} designing service networks to meet demand while minimizing cost. In this work we focused on large-scale tactical planning that is restricted to deterministic models for the sake of computational tractability. Even though \greta{the} estimate of \greta{the} periodic demand is a central input to such models, the associated estimation problem has not been studied in the literature. In this paper we addressed this gap: We formally introduced the periodic demand estimation problem and we proposed a methodology that proceeds in two steps. The first step consists in using a time series forecasting model to predict demand for each period in the tactical planning horizon. The second step defines periodic demand as a solution to a multilevel mathematical program that explicitly connects the estimation problem to the tactical planning problem of interest. This allows to estimate periodic demand such that the costs are minimized. Since the origin-destination demand matrices typically are unbalanced, this can be of importance as the cost of forecast errors is not evenly distributed across commodities. 

We reported results for a real large-scale application at the Canadian National Railway Company. The results clearly showed the importance of the periodic demand estimation problem when compared to the approach commonly used in practice. The latter consists in averaging the time series forecasts over the tactical planning horizon. Compared to this practice, the results showed that using another estimate can lead to a substantial reduction in cost. 
As expected, the results also showed that the time series forecasting problem is difficult and the forecast errors hence are relatively large.
Nevertheless, the periodic demand estimates that resulted from the proposed methodology still led to costs that were comparable \ef{to}, or even better, than those obtained by using the average demand baseline computed on \emph{perfect information} (i.e., no forecast error). Moreover, the costs were substantially reduced compared to averaging the forecasts. 

In terms of exposition, we chose to limit the methodology to the MCND formulation. However, the methodology \ef{may apply} to other cyclic network design formulations. Similarly, adaptation of the tactical plan in each period (\textbf{wMCND} and \textbf{wBP} formulations) can also be represented differently from this paper. The methodology hinges on the separation between the design variables that are fixed for all periods in the tactical planning horizon while the flow decisions are not. The adaptation of the flow decisions serves as a proxy for operational costs.

 The work reported in this paper constituted a first step in addressing the periodic demand estimation problem that hitherto has been overlooked in the literature. We showed that adequately addressing it can lead to important cost reductions. Given that we introduced a new problem, it opens up a number of directions for future research. \ef{Two direct extensions consist in} improving the time series forecasts (step one in the methodology)\ef{, and}
 extending the feasible set of periodic demand values to more general mappings\ef{. The latter requires an effective solution approach.} 
 
\ef{Another promising avenue for future research consists in formally linking the periodic demand estimation problem to other related research areas. Here we identify two such areas. First, stochastic programming. We assume a deterministic and cyclic formulation for computational tractability. However, computing power improves fast and a stochastic formulation is a promising avenue, for example, taking into account the uncertainty around the point forecasts (error distributions or prediction intervals).
%scenario generation for sample average approximation in stochastic programming. Our approach differs from the latter as we estimate the periodic demand rather than a tactical plan. The third level of our formulation can, however, be viewed as the generation of a finite number of scenarios. We could draw inspiration from the scenario generation in stochastic programming to improve our paradigm, for example, taking into account forecast error distributions (or prediction intervals). 
% Our approach may be viewed as the generation of one single representative scenario. However, it may be possible to extend the approach to generate multiple high-quality scenarios, for example, taking into account forecast error distributions (or prediction intervals). 
The second area of research is end-to-end learning that focuses on incorporating the value of a prediction for a downstream decision-making problem already in the training / estimation \citep[see, e.g.,][]{Bengio97,FerberEtAl20,kotaryEtAl21}. Such criterion stand in contrast to traditional prediction ones (e.g., minimizing squared errors). While there is a recent surge in the related literature, none of the existing methods can tackle formulations of the type we consider here. Namely, MILP formulations where predictions occur in the constraints.}

\section*{Acknowledgments}
This research was funded by the Canadian National Railway Company (CN) Chair in Optimization of Railway Operations at Université de Montréal and a Collaborative Research and Development Grant from the Natural Sciences and Engineering Research Council of Canada (CRD-477938-14). Moreover, we gratefully acknowledge the close collaboration with personnel from different divisions of CN. 
We would also like to thank the \efsec{five} anonymous reviewers as well as Frederic Semet and Jonathan Jalbert who provided valuable comments that helped us improve the manuscript.

\bibliographystyle{plainnat_custom}
\bibliography{all-ref}

\section*{Appendix}

\subsection*{Input features of neural networks}

\paragraph{Input Layer 1} It is dedicated to external data features such as weather data and temporal context, for all models. We use real observed weather data described in Section~\ref{subsec:data}, to assess their potential in an optimistic setting in regards to their accuracy. At prediction time, we would then rely on weather forecasts.  

All models contain two features as temporal context, that is the week number of the forecasted week and the month number of the Monday of the forecasted week. Weather features, for models that include them, consist in the average daily temperature, the accumulated snow (cm) and the accumulated precipitations (mm) of the forecasted week for the main 17 terminals in the network.

\paragraph{Input Layer 2} Features in Input Layer 2 are lagged observed or forecasted (when doing inferences) demand of commodities predicted by the model. The number of lags was found through hyperparameters optimization, and is given in Table~\ref{table:hyperparams} below.

\subsection*{Hyper-parameters of neural networks}
\begin{table}[H]
\centering
\begin{tabular}{lcccccccccc}
\textbf{Model} & \textbf{Lags} & \textbf{\begin{tabular}[c]{@{}c@{}}Weather \\ Features\end{tabular}} & \textbf{\begin{tabular}[c]{@{}c@{}}Hidden \\ layers\end{tabular}} & \textbf{\begin{tabular}[c]{@{}c@{}}Size \\Hidden \\ Layers\end{tabular}}  & \textbf{Dropout} & \textbf{\begin{tabular}[c]{@{}c@{}}Learning \\ Rate\end{tabular}}  \\ \hline
RNN & 3 & NO & 1 & 700  &  0.25 & 0.1  \\
RNN-W & 4 & YES & 1 & 260  &  0.12 & 0.1 \\
RNN-W-SPLIT1 & 3 & YES & 3 & 460  &  0.10 & 0.1 \\
RNN-W-SPLIT2 & 8 & YES & 2 & 220  &  0.23 & 0.1 \\
FFNN & 3 & NO & 3 & 540 & 0.14 & 0.01 \\
FFNN-W & 4 & YES & 3 & 600  &  0.11 & 0.1 \\
FFNN-W-SPLIT1 & 3 & YES & 2 & 700  &  0.06 & 0.01 \\
FFNN-W-SPLIT2 & 5 & YES & 3 & 580  &  0.15 & 0.1 
\end{tabular}
\caption{Table of hyperparameters of the neural networks}
\label{table:hyperparams}
\end{table}

\end{document}